\documentclass[lettersize,journal]{IEEEtran}
\usepackage{amsmath,amsfonts}
\usepackage{algorithmicx}
\usepackage{algorithm}
\usepackage{algpseudocode}
\usepackage{array}
\usepackage[caption=false,font=normalsize,labelfont=sf,textfont=sf]{subfig}
\usepackage{textcomp}
\usepackage{stfloats}
\usepackage{url}
\usepackage{verbatim}
\usepackage{graphicx}
\usepackage{cite}
\usepackage{multirow}
\usepackage{tabularx,booktabs,caption}
\usepackage{arydshln}
\begin{document}

\title{Federated Learning for Traffic Flow Prediction with Synthetic Data Augmentation}

\author{Fermin Orozco, Hongkai Wen, Pedro Porto Buarque de Gus\~{m}ao, Johan Wahlström, Man Luo}

% The paper headers
\markboth{Journal of \LaTeX\ Class Files,~Vol.~14, No.~8, August~2021}%
{Shell \MakeLowercase{\textit{et al.}}: A Sample Article Using IEEEtran.cls for IEEE Journals}

\maketitle

\begin{abstract}
Deep-learning based traffic prediction models require vast amounts of data to learn embedded spatial and temporal dependencies. The inherent privacy and commercial sensitivity of such data has encouraged a shift towards decentralized data-driven methods, such as Federated Learning (FL). Under a traditional Machine Learning paradigm, traffic flow prediction models can capture spatial and temporal relationships within centralised data. In reality, traffic data is likely distributed across separate data silos owned by multiple stakeholders. In this work, a cross-silo FL setting is motivated to facilitate stakeholder collaboration for optimal traffic flow prediction applications. This work introduces an FL framework, referred to as FedTPS, to augment each client's local dataset with generated synthetic data by training a diffusion-based traffic flow generation model through FL. The proposed framework is evaluated on two large scale real world ride-sharing datasets and assessed against various generative and spatio-temporal prediction models, including a novel prediction model we introduce which leverages Temporal and Graph Attention mechanisms to learn embedded Spatio-Temporal dependencies. Experimental results show that FedTPS outperforms several FL baselines with respect to global model performance. 
\end{abstract}


\section{Introduction}
\IEEEPARstart{M}{odern} Intelligent Transport Systems (ITS) have increasing levels of connectivity enabling data exchange between users and infrastructure; it is projected that by 2026, all vehicles on UK roads will have some degree of connectivity \cite{ConnectedNumber2026}. This data can be leveraged to develop traffic prediction models, which can enhance the safety and productivity of users, as well as deliver significant social, environmental, and economic benefits to a city’s inhabitants. Deep learning-based models have demonstrated exceptional accuracy in their ability to model spatio-temporal relationships, and predict future traffic states \cite{STGCN, DCRNN, GWN, TAU, GRU}.

The majority of previous works towards spatio-temporal prediction aim to train a model from a centralised perspective where the collective data is available on a single device. Centralising city-wide ITS data in real-world settings is a difficult task; ITS represent a partnership between data-owning stakeholders such as local government authorities, ride-sharing companies, and micro-mobility organisations. Sharing journey data has implications on user privacy (GDPR), as well as commercial implications. The development of frameworks for decentralised data-driven approaches, which eliminate the need to centralize private and commercially sensitive information, is therefore critically important within the ITS sector for fostering stakeholder collaboration.

\begin{figure}[h!]
\vspace{-3mm}
  \centering
  \includegraphics[width=\linewidth]{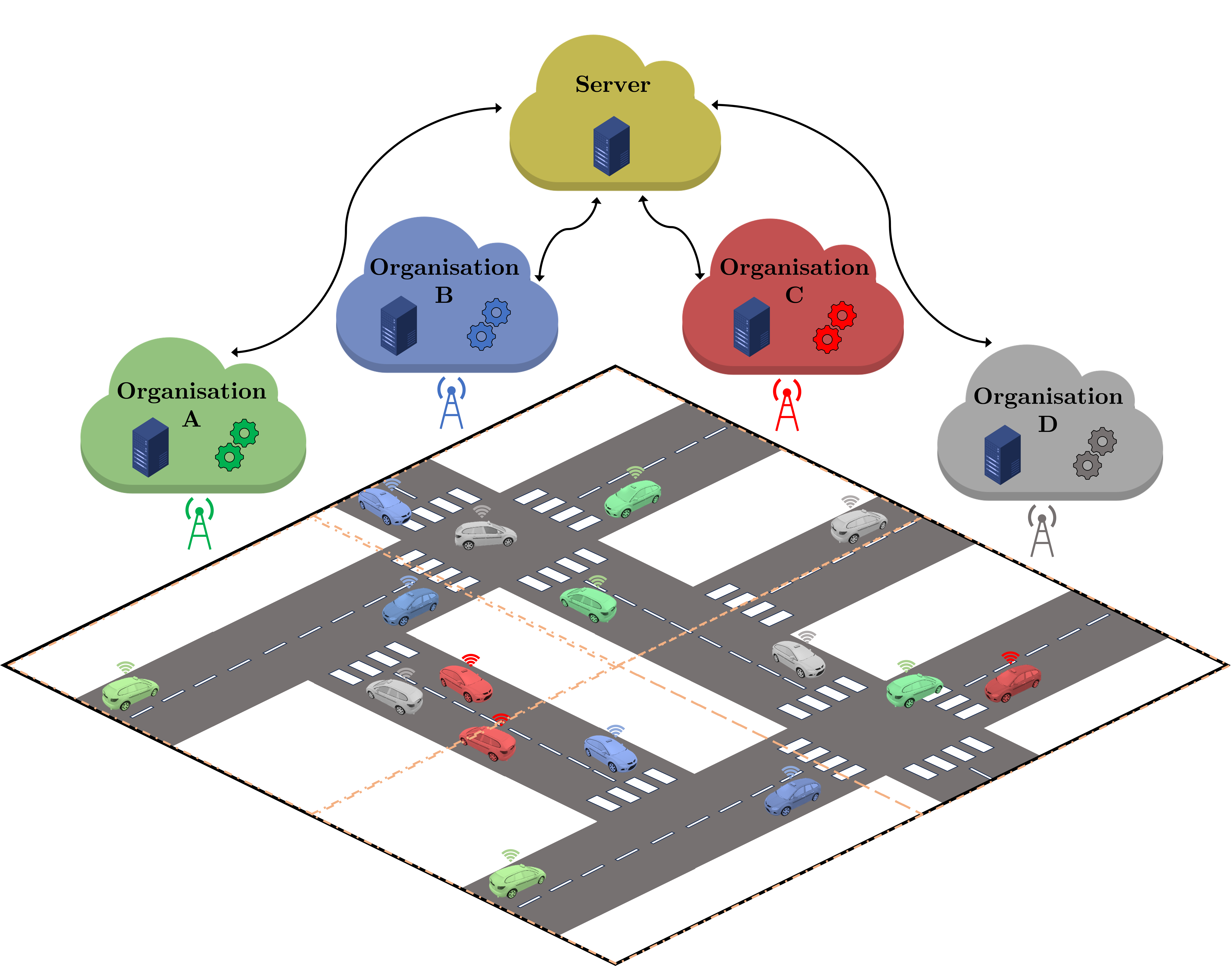}
    \caption{Federated framework for the traffic flow prediction task within an ITS. Each organisation collects data from their respective vehicle fleet, and through FL, collaboratively train a model to estimate traffic flow between regions (shown as orange grid cells). 
    \label{fig:Front_paper_figure}
}%
\vspace{-3mm}
\end{figure}

Federated Learning (FL) \cite{FedAvg} provides a solution for training traffic prediction models collaboratively by decentralising the training process of a traffic prediction model. Such a framework is detailed further in Figure \ref{fig:Front_paper_figure}. Although recent works in FL demonstrate its capacity to train a global model which learns spatio-temporal dependencies across distributed data, these works do not directly address the issue of data heterogeneity \cite{FedTDP}, or instead focus on the client selection strategy or parameter aggregation function \cite{CNFGNN, FedSTN, FedGRU, FastGNN, FedDA, GNN_CTFL, FedACGN}.

To tackle the issue of data heterogeneity and imbalanced data partitions across clients, data sharing mechanisms for FL applications in highly heterogeneous settings were explored in \cite{nonIIDFL} wherein a small partition of real data is distributed to clients to augment their local datasets. However, such an approach which shares real data between clients may violate privacy requirements. 

Research within FL for image classification tasks have recently looked at training generative models to augment local client datasets without sharing data that would compromise data privacy \cite{2022GANSynthFL, FedGP, FedSyn}. For applications beyond image classification where data contains spatial and temporal dependencies and heterogeneity is principally due to quantity and feature-based skews, limited work exists on the benefits of augmenting FL client datasets using synthetic data.

Emerging research in the field of generative models have demonstrated the exceptional capacity for models to generate synthetic data that capture spatio-temporal interdependencies of traffic data \cite{DiffTraj}. Leveraging recent works in FL as well as in the field of generative models for spatio-temporal data generation, we present a framework for Federated Traffic Prediction with Synthetic data augmentation (FedTPS). This framework first trains a federated diffusion-based model for traffic flow data generation and generates a synthetic dataset developed from the learned global data distribution. A traffic flow prediction model is then trained through FL, wherein each client's local dataset has been augmented with a generated synthetic dataset.

In real-world implementations of traffic prediction models, the data owned by separate clients may contain inherent skews, contributing to data heterogeneity across client datasets. The proposed FedTPS framework directly inhibits the negative effects of data heterogeneity when training a global model through FL, while also increasing the amount of data in each client's local dataset. To the best of our knowledge, this work is the first to explore the benefits of augmenting the federated framework with synthetic data in traffic prediction applications. 

The main contributions of our work are as follows:
\begin{itemize}
    \item We present a novel traffic prediction model, the Graph Attention and Temporal Attention Unit (GATAU), which builds upon the existing spatio-temporal model, the Temporal Attention Unit (TAU) \cite{TAU}, to also leverage Graph Attention mechanisms. 
    \item We propose a novel FL framework, FedTPS, which trains a federated generative model to produce a synthetic traffic flow dataset. This synthetic data is used to augment client datasets for traffic flow prediction, and improve the predictive performance of the global FL model. 
    \item We assess the performance of several traffic flow prediction models trained under an FL setting. Our experiments are validated on two real-world datasets to analyse the impact of data augmentation on traffic flow prediction model performance, and demonstrate improved results relative to various FL frameworks designed to address data heterogeneity across data-silos.
\end{itemize}

\section{Related Work}
\subsection{Federated Learning}
The distributed nature of the data in an FL setting has a detrimental effect on the accuracy of the model relative to centralized training paradigms due to the fact that the distribution of data among clients is non-Independent and Identically Distributed (non-IID). In cross-silo FL settings, each client may represent a data center, owned by a particular organisation, which stores large amounts of data aggregated from many devices. FL has been extended to optimise the training of the global model \cite{FedOpt}, as well as to improve its applicability in non-IID settings \cite{FedProx}. Yet such methods focus on optimising the FL framework rather than address the data heterogeneity issues directly. 

Existing data augmentation approaches for FL focus on image classification tasks, and typically rely on Generative Adversarial Networks (GANs) \cite{FedGP, 2022GANSynthFL, FedSyn}, or on sharing mean latent representations of client data \cite{FedMix, CVPR_SyntheticData_FL_VAE}. Unlike image generation applications, traffic data generation requires modelling the dynamics of data over both spatial and temporal dimensions. Differentially-private representations of spatio-temporal data can be produced by altering real data \cite{GeogMasking, ConfidentialityGeocoded}, or mixing partial aspects of the data \cite{TrajMixing}. These processes affect the spatio-temporal characteristics of the data itself, however. Recent work into diffusion-based models for spatio-temporal data generation demonstrate their capacity to generate high-fidelity synthetic trajectories which capture the spatio-temporal characteristics of sample traffic data, while ensuring the privacy of the real data.

Existing work on FL applied for traffic prediction implement novel client selection strategies to cluster clients with similar spatial characteristics \cite{GNN_CTFL, FedDA, CNFGNN}. Other such works look to adapt the aggregation procedure of the FL setting \cite{FastGNN, FedSTN, FedACGN}. FedTDP \cite{FedTDP} explores the data heterogeneity issue which may be found in real-world traffic data silos, yet does not propose a method for dealing with the inherent data heterogeneity across client data silos. 

\subsection{Traffic Flow Modelling}

To extract temporal dependencies, deep-learning based models typically employ either Recurrent Neural Networks (RNNs) based architectures \cite{DCRNN, FedGRU}, temporal Convolutional Neural Networks (CNNs) which convolve around the time dimension of traffic flow data \cite{STGCN, GWN}, or temporal attention mechanisms to capture the evolution of the data over time \cite{TAU}. 

Spatial relationships embedded within data can be captured through CNNs which convolve around the spatial dimension of data \cite{TAU}. Traffic data can also be represented as graph networks, enabling Graph Convolutional Networks (GCNs) to capture spatial relationships within the data. GCN operations are dependent on the graph structure, which requires expert contextual knowledge to capture inter-region relationships. Graph Attention Networks (GATs) \cite{GAT} implement a self-attention mechanism to learn the relationship between nodes.

To capture spatio-temporal dependencies within data, models such as STGCN \cite{STGCN} implement GCNs with temporal CNNs to capture spatio-temporal dependencies within data, while DCRNN \cite{DCRNN} leverages GCNs with RNN-based models. GraphWaveNet \cite{GWN} also utilises temporal CNNs and GCNs, however it learns an adjacency matrix for the graph structured data. The Temporal Attention Unit (TAU) \cite{TAU} utilises CNNs and temporal Attention mechanisms to capture both spatial and temporal relationships within data.

Generative approaches have been explored for synthesising realistic and diverse spatio-temporal data \cite{TrajGAN, CHEN2021332}. Conditional Generative Adversarial Networks (cGANs) \cite{vanilla_CGAN} extend the standard GAN architecture by conditioning both the generator and discriminator on auxiliary variables (e.g. traffic density, or environmental context) to enhance control over the generation process. Similarly, Conditional Variational Autoencoders (cVAEs) \cite{vanilla_cVAE} integrate conditional information into the latent space of the VAE framework. DiffTraj \cite{DiffTraj} is a conditional diffusion-based trajectory generation model which leverages Traj-UNet, a network based on the UNet architecture \cite{Unet}, to model the noise transitions at each time step in the diffusion process. The model incorporates conditional information, embedded by a Wide and Deep network \cite{Wide_and_Deep}, to enhance the generative ability of the model. 

\section{Problem Definition}

\textit{\text{\textbf{Definition 1 (Traffic Flow Generation):}} 
To emulate the setting of various organisations each with ownership over separate data silos, the global trajectory dataset, $X_{\textnormal{glob}}$, is partitioned into $C$ clients such that $X_{\textnormal{glob}} = (X_1 \cup X_2 \cup ... \cup X_{C})$, where $X_c$ is client $c$'s trajectory dataset. Traffic flow data is obtained from $X_c$ by discretising the city into $N$ regions, and defining $q_c^t \in \mathbb{R}^{N}$ as client $c$'s traffic inflow over $N$ regions at time step $t$. The objective of data generation is, given a set of traffic inflows, $Q_c = \{ q_c^1, q_c^2, \ldots, q_c^m \}$, learn a model, $G$, which can generate a set of synthetic inflows, $Q_{\textnormal{synth}} = \{ q_{\textnormal{synth}}^1, q_{\textnormal{synth}}^2, \ldots, q_{\textnormal{synth}}^m \}$, where the synthetic data retains the real data's spatial-temporal characteristics. The federated task is to train model G over $X_{\textnormal{glob}}$ through FL using the distributed client datasets.}

\textit{\text{\textbf{Definition 2 (Traffic Flow Prediction):}} 
 Given historical traffic flow $ Q = (q_1, q_2, ... , q_{T}) \in \mathbb{R}^{N \times T}$, train a model to predict the traffic inflow of all regions at future time step $T + \tau$, hence predict $q_{T+\tau}$. Based on the partitioned trajectory data $X_{\textnormal{glob}}$, client $c$ can derive its regional traffic inflow data $Q_c$, and can also augment its own local dataset using the synthetic dataset $Q_{\textnormal{synth}}$. The federated traffic flow prediction task is then to collaboratively train a global prediction model using the distributed client datasets with synthetic data augmentation, through FL. }

\section{Methodology}
This work aims to address a real-world setting of multiple organisations, each with their independent dataset of vehicle trajectories, collaborating through FL to train a trajectory generation model and subsequent traffic flow prediction model. To replicate this scenario, a large-scale, real-world trajectory dataset is segmented into subsets representing local client datasets in a cross-silo FL setting. Within this FL setting, a server acts as a secure, trusted central system which coordinates federated training between clients.

The proposed FedTPS framework first aims to train a generative model through the partitioned vehicle trajectory datasets to generate a synthetic dataset from the global distribution of data. The second step aims to train a traffic flow prediction model using each client's traffic flow data, and augment each client's data with the synthetic generated data. This will reduce the effects of data heterogeneity present across the client datasets, and increase the amount of training data available to clients. 

\subsection{Federated Generative Model}
As a probabilistic generative model, a diffusion model aims to generate synthetic data by modeling two sequential processes: 1) A series of forward processes which gradually injects noise into the data, and 2) The reverse of these processes, which learn to capture the original data distribution from a noisier version \cite{OG_Diffusion}. Recent works have applied diffusion-based models for high-fidelity data generation in various domains \cite{DiffuSeq, Imagen, Geodiff}. The forward process is defined as a Markov chain with $F$ Gaussian transitions that map the real data $x_0 \sim g(x_0)$ to $x_F$, a latent variable with the same dimensionality as $x_0$, according to

\begin{equation}
\label{Diffusion_Forward_1}
    g(x_1, ..., x_F | x_0) = \prod^{F}_{f=1} g(x_f | x_{f-1}),
\end{equation}
\begin{equation}
\label{Diffusion_Forward_2}
    g(x_f | x_{f-1}) = \mathcal{N} (x_f;\sqrt{1- \beta_{f}} x_{f-1}, \beta_f \mathbf{I}),
\end{equation}

\noindent where $\beta_f$ denotes the variable variances which control the Gaussian noise. Importantly, the forward process allows sampling $x_f$ in closed form from $x_f = \sqrt{\bar{\alpha_f}}x_0 + \sqrt{1-\bar{\alpha_f}}\epsilon$, where $\epsilon \sim N(0,\mathbf{I})$ and $\bar{\alpha_f} = \prod^{f}_{i=1} (1-\beta_{i})$. The reverse process can be parameterized as a Markov chain with learned Gaussian transitions
\begin{equation}
\label{Diffusion_Reverse}
    p_{\theta}(x_0, ..., x_{f-1}|x_{F}) = p(x_{F})\prod^{F}_{f=1}p_{\theta}(x_{f-1}|x_f),
\end{equation}
\begin{equation}
\label{Diffusion_Reverse_2}
    p_{\theta}(x_{f-1}|x_{f}) = \mathcal{N}(x_{f-1}; \mu_{\theta}(x_f, f), \sigma_{\theta}(x_f, f)^2 \mathbf{I}),
\end{equation}

\noindent where $x_{F} \sim \mathcal{N}(0,1)$, and where $\mu_{\theta}(x_f, f)$ and $\sigma_{\theta}(x_f, f) $ are the mean and variance, parameterized by $\theta$. The objective of the diffusion model is to minimise the error between the Gaussian noise $\epsilon$ and the predicted noise level $\epsilon_{\theta}(x_f, f)$, over its training set. For a data sample $x_0$, the client loss $L_c$ is given by 

\begin{equation}
\label{Local_Diffusion_Training}
    \mathcal{L}_c(\theta) := \mathbb{E}_{x_0\in X_c, \epsilon, f} \left[ || \epsilon - \epsilon_{\theta} (x_{f}, f) ||^2 \right].
\end{equation}

The UNet architecture \cite{Unet} has proved highly effective in generative modeling tasks; the fusion of multiple scales allows a diffusion model built on UNet to robustly navigate noise removal in each denoising step of the diffusion process. UNet employs a contracting encoder, which applies convolutional operations to progressively reduce spatial resolution and extract salient features, followed by an expanding decoder that upsamples these encoded representations to restore the original spatial dimensions. Skip connections link corresponding encoder and decoder layers to retain spatial information from multiple scales that may otherwise be lost during downsampling; as the decoder upsamples the encoded input, the skip connections provide finer-scale information from the earlier encoded layers.

For the trajectory generation task, Zhu et al. \cite{DiffTraj} introduce an attention-driven transition module, and demonstrate its ability to bolster the spatio-temporal modeling capacity of diffusion-based methods. Akin to their DiffTraj model, our Traffic Flow Diffusion model (TFDiff) utilises the UNet architecture, with an intermediary attention block.

Conditional diffusion models enable additional information to be embedded into the reverse denoising process as conditional variables into Eq. \eqref{Diffusion_Reverse_2}. TFDiff is conditioned on information regarding the time interval, the day of the week, the maximum traffic flow at any region within the time interval, and the region index with the maximum traffic flow. 

\subsubsection{Federated Framework}

In a scenario where the global trajectory dataset is segmented across data silos from multiple organisations, a generative model cannot be trained under a traditional framework since the data is not centralised.

\begin{figure*}[h]
\vspace{-5mm}
  \centering
  \includegraphics[width=0.95\textwidth]{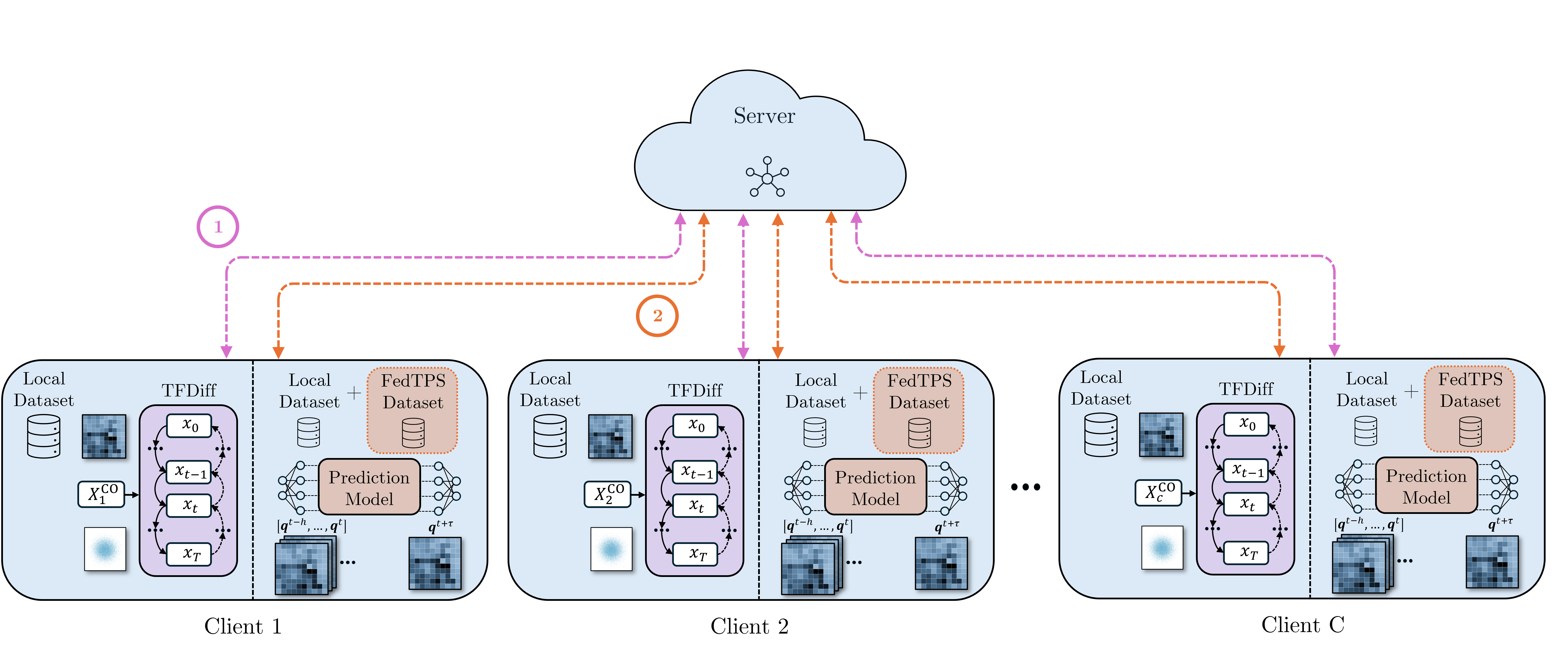}
  \vspace{-3mm}
  \caption{FedTPS Framework for training a federated generative model and subsequent traffic flow prediction model. The first stage, shown as the purple dashed line, trains the federated diffusion model to develop the FedTPS synthetic dataset. The second stage, shown as the orange dashed line, trains the federated traffic flow prediction model with the client's local dataset, as well as the synthetic data. $X^{\textnormal{CO}}_c$ represents client $c$'s conditional observations required for the TFDiff model.}
  \label{fig:FedTPS_Framework}
\vspace{-5mm}
\end{figure*}

Under a federated setting, the updated parameters from each client's local training on the global model are communicated to a central server which then aggregates the results to update the global model. FedAvg aggregates the parameters based on the number of data samples in a client's partition. The objective of the federated TFDiff model is to minimise the loss of the model over the cross-client datasets, and can be formalised as 

\begin{equation}
\label{Global_TFDiff}
    \underset{\theta}{\arg \min}\;L(\theta) := \sum^{C}_{c=1} \frac{|Q_{c}|}{|Q_{\textnormal{glob}}|}L_c(\theta)
\end{equation}

\noindent where $|Q_{c}|$ and $|Q_{\textnormal{glob}}|$ denote the number of traffic flows in the dataset of client $c$ and in the global set of all clients, respectively. The process of training a federated Diff model is illustrated in step one of Figure \ref{fig:FedTPS_Framework}, and is detailed further in Algorithm \ref{alg:TFDiff_FL}. Once the global TFDiff model has been trained, it is shared with all clients, who then use it to generate the synthetic datasets to augment their respective local datasets.

\begin{algorithm}
    \caption{Training the TFDiff model through FL using FedAvg.}\label{alg:TFDiff_FL}
    \begin{algorithmic}[1]
        \State Server initialises global model $\theta$
        \State Server selects all clients for participation, $c \in C$
        \For {$\textnormal{global round, } r_{\textnormal{glob}}\in  \{1,\dots, R_{\textnormal{glob}}\}$}
        \For {$\textnormal{each client } c \in C \textnormal{ in parallel}$}
        \State Distribute global model to clients, $\theta \rightarrow c$
        \For {local round $r_{loc} \in (1, ..., R_{loc})$}
        \State Sample $x_0 \sim g(X_c)$
        \State Sample $f \sim \textnormal{Uniform} \{1, ..., F\}$
        \State Sample $\epsilon \sim \mathcal{N}(0, \mathbf{I})$
        \State Local updates $\theta^{c}_{r_{loc}+1} \leftarrow \theta^{c}_{r_{loc}} - \eta \nabla_{\theta}\mathcal{L}_c$
        \EndFor
        \State Global updates $\theta_{r_{\textnormal{glob}}+1} \leftarrow \sum_{c=1}^C \frac{|Q_c|}{|Q_{\textnormal{glob}}|}\theta_{r_{\textnormal{glob}}}^c$
        \EndFor
        \EndFor
    \end{algorithmic}
\end{algorithm}

The conditional data derived from each client's private dataset, as well as the distributed global TFDiff model, are used by each client to generate their synthetic dataset. Synthetic data is generated for a duration of 10 days. Generating additional synthetic data did not yield significant improvement in the global federated traffic prediction model.

\subsection{Federated Traffic Flow Prediction Task}

Using each client's local individual trajectory dataset $X_c$, the regional inflow dataset, $Q_c$, can be constructed, where $Q_c \in \mathbb{R}^{T \times N}$, $T$ is the number of regional traffic flow interval samples, and $N$ is the number of regions within the city. Each client can augment their local dataset with the synthetic regional inflow dataset $Q_{\textnormal{synth}}$ generated from each client using the TFDiff model for inference. Step two in Figure \ref{fig:FedTPS_Framework} illustrates the structure of the FedTPS framework for the traffic prediction task.

The local objective of a traffic flow prediction model is denoted by

\begin{equation}
\label{eq:Local_TFP_objective}
    f_c(\omega) = \frac{1}{|Q_{c}|}\sum_{t}(\mathcal{L}(\mathcal{F}_c(\omega; q_{t-h:t}), q_{t+\tau})
\end{equation}

\noindent where the model is parameterised by $\omega$, $h$ denotes the number of previous traffic flow data samples fed to the model, and $\tau$ is the number of time steps in the future for which we will be predicting traffic flow. The global objective is therefore 

\begin{equation}
    \label{eq:Global_TFP_objective}
     \textnormal{argmin}f(\omega) := \sum^C_{c=1}\frac{|Q_c|}{|Q_{\textnormal{glob}}|}f_c(\omega)
\end{equation}

\noindent where $|Q_{c}|$ and $|Q_{\textnormal{glob}}|$ denote the number of regional inflow data samples in the dataset of client $c$, and in the global set of all clients, respectively.

\begin{algorithm}
    \caption{Training a global traffic flow prediction model through FedTPS.}\label{alg:FL_TrafficFlowPrediction}
    \begin{algorithmic}[1]
        \State Server generates synthetic data, $X_{\textnormal{synth}}$, from $\theta$
        \State Server initialises global model $\omega$
        \State Server selects all clients for participation, $c \in C$
        \State Server disseminates synthetic data, $X_{\textnormal{synth}} \rightarrow C$ 
        \For {$\textnormal{global round, } r_{\textnormal{glob}}\in  \{1,\dots, R_{\textnormal{glob}}\}$}
        \For {$\textnormal{each client } c \in C \textnormal{ in parallel}$}
        \State Distribute global model to clients, $\omega \rightarrow c$
        \For {local round $r_{loc} \in (1, ..., R_{loc})$}
        \State Local updates $f^{c}_{r_{loc}+1}(\omega) \leftarrow f^{c}_{r_{loc}}( \omega) - \eta \nabla_{\omega}\mathcal{L}_c$
        \EndFor
        \State Global updates $\omega_{r_{\textnormal{glob}}+1} \leftarrow \sum_{c=1}^C \frac{|Q_c|}{|Q_{\textnormal{glob}}|}f(\omega)_{r_{\textnormal{glob}}}^c$
        \EndFor
        \EndFor
    \end{algorithmic}
\end{algorithm}
  \vspace{-3mm}
  
\subsection{Proposed Traffic Flow Prediction Model}

The Temporal Attention Unit was introduced by Tan et al \cite{TAU} and poses temporal attention as a combination of statical attention, which occurs within a data sample, and dynamical attention, which occurs between the frames of the data samples. The statical attention can be captured by performing depth-wise convolution, followed by a dilated depth-wise convolution, and finally a 2D convolution. The depth-wise convolution layers learn separate  kernel filters for each of the channels in the encoded representation of the input. The dynamical attention is composed of a 2D average pooling layer, followed by a fully-connected layer. The final temporal attention is a result of the product between the statical and dynamical attention components. 

\begin{figure}[h]
\vspace{-2mm}
    \centering
    \includegraphics[width=5cm]{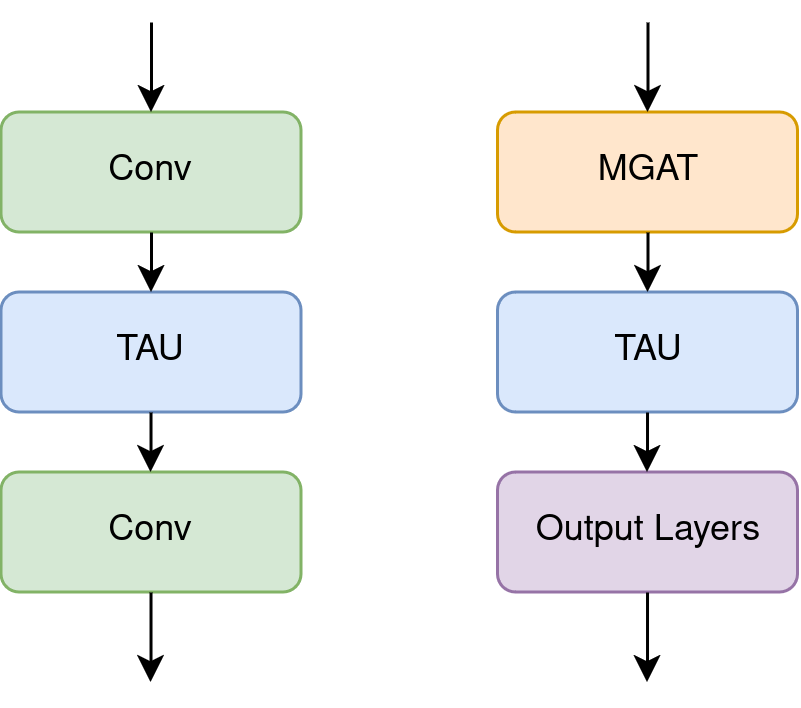}
    \caption{General model architecture comparison of TAU and GATAU.}
    \label{fig:Architecture_Comparison}
    \vspace{-4mm}
\end{figure}

An Encoder-Decoder architecture model architecture is presented in this work, wherein the Encoder and Decoder represent stacked 2D convolutional layers that encode a spatial representation of the data. To augment this Encoder-Decoder architecture for the task of traffic prediction, we can model the regional traffic flow data as a sequence of graphs, and employ the Graph Attention model \cite{GAT} with multi-head attention, to learn multi-channel relationships between regions in the city. Figure \ref{fig:Architecture_Comparison} compares the model architecture between TAU and GATAU. 

\begin{figure}[h]
\vspace{-2mm}
  \centering
  \includegraphics[width=\linewidth]{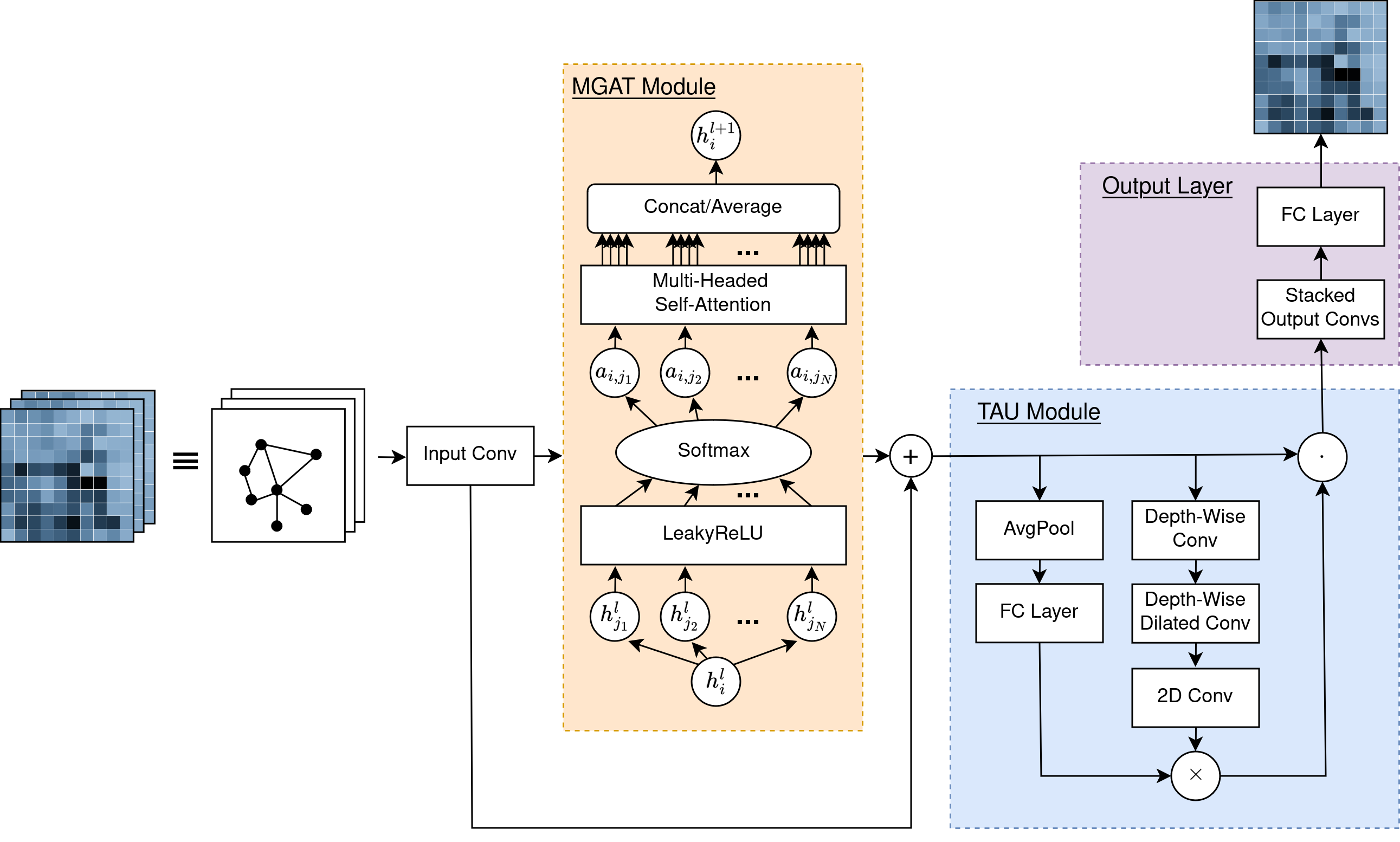}
  \caption{Architecture of the proposed GATAU model.}
  \label{Proposed_Model}
\end{figure}

Thus, we propose the Graph Attention with Temporal Attention Unit (GATAU). Figure \ref{Proposed_Model} illustrates the GATAU module in further detail.

Our input data $Q \in \mathbb{R}^{N \times T}$, where $N$ is the number of nodes and $T$ is the length of the input sequence, is first processed by a 2D convolutional layer to map the input into a higher dimension feature space. This output is fed into the Multi-Head Graph Attention (MGAT) module. Within the MGAT, attention coefficients are calculated which give a measure of the importance of relationships between nodes. The input into MGAT are node features, $h = \{ \vec{h}_1, \vec{h}_2, ..., \vec{h}_N\} \in \mathbb{R}^{T\times F}$ where $F$ is the input feature size. Attention coefficients for layer $l$ are calculated by

\begin{equation}
\label{eq:GAT_Attention_1}
    e_{ij}^{l}= \textnormal{LeakyReLU} \left( \vec{a}^{l^{T}} \left[W^{l}   \vec{h}^{l}_{i} || W^{l} \vec{h}^{l}_{j}\right] \right)
\end{equation}

\noindent where $W \in \mathbb{R}^{F' \times F}$  is a learnable weight matrix, $F$ is the feature dimension of the input, $F'$ is the output feature dimension which can be of different cardinality to $F$, and $a$ is a function which computes the attention score from the concatenation of linearly transformed embedding of neighbouring nodes. We further apply a LeakyReLU activation for the attention coefficients. A fully connected linear single layer is used to represent function $a$. Then we normalise the attention scores across all connections for a given node through

\begin{equation}
\label{eq:GAT_Attention_2}
    a_{ij}^{l} = \frac{\textnormal{exp} (e^{l}_{ij})} {\sum_{k \in \mathcal{N}_i} \textnormal{exp}(e_{ik}^{l})}
\end{equation}

\noindent where $\mathcal{N}_i$ denotes the node connections of node $i$. The output features for every node are calculated as a linear combination of the normalised attention coefficients from Equation \ref{eq:GAT_Attention_2} to give

\begin{equation}
    \label{eq:GAT_Attention_3}
    h_{i}^{l+1} = \sigma \left( \sum_{j \in \mathcal{N}_i} \alpha_{ij}W^{k} \vec{h}_j \right)
\end{equation}

\noindent where $\sigma$ is an ELU activation function. To enhance the stability of the model during training, a multi-head mechanism of attention is utilised for the graph attention module, wherein each attention head trains its own parameters. As per the original work which averages the multi-head outputs at the final layer \cite{GAT}, our model also averages the multi-head outputs after the single-layer MGAT module. The output from the MGAT module is of shape $[N, T, F']$.

A residual connection is used to fuse the output of the initial convolution layer with the output from the MGAT module. The fused hidden layer is then reshaped into $[N, F', H, W]$ format to be processed by the TAU module, where $H$ and $W$ are the dimensions of the rectangular grid of the city. Our output layer is composed of a convolutional layer, and fully connected layer, which reduce the dimensionality of our processed input's feature space to match our target shape, which is $Y \in \mathbb{R}^{N}$.

\section{Experiments}

\subsection{Dataset}
The experiments in this study utilise the ride-sharing dataset published by Didi Chuxing GAIA Initiative, spanning the cities of Chengdu and Xi'an in China during November 2016. Chengdu has a population of 21 million and features a ring-road-based road network, consistent traffic patterns influenced by stable weather, little prominent tourism variation, and broader spatial traffic distribution due to urban sprawl. In contrast, Xi'an has nearly 13 million residents and features a grid-based road network, greater traffic variability due to weather and tourism, and a dense historic core that concentrates traffic patterns. The different characteristics of these two datasets provide a diverse study to evaluate FedTPS's ability to enhance traffic flow prediction models in different settings. Table \ref{tab:Chengdu_Dataset} contains details of the datasets. 

\begin{figure}[h!]
    \centering
    \subfloat{{\includegraphics[width=4.25cm]{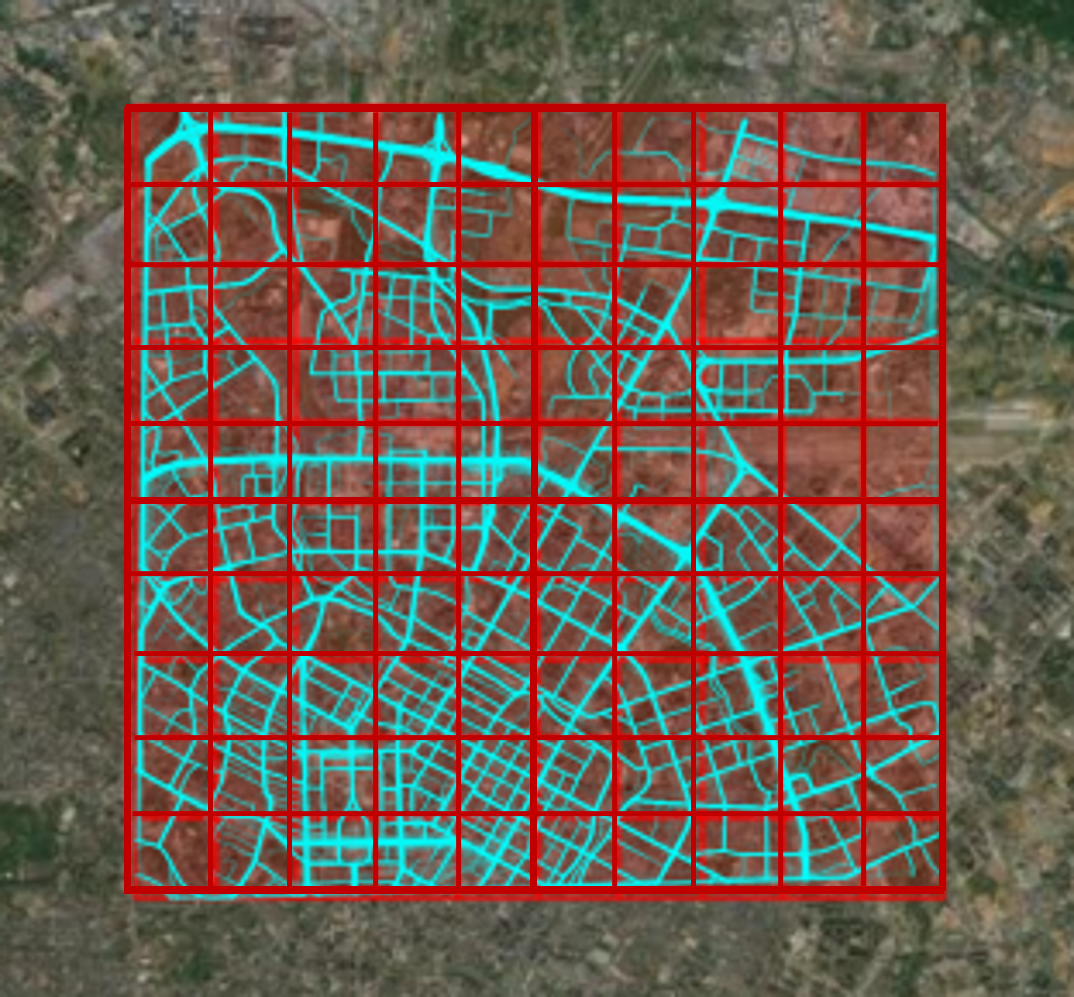} }}%
    \hspace{0.35cm}%
    \subfloat{{\includegraphics[width=4cm]{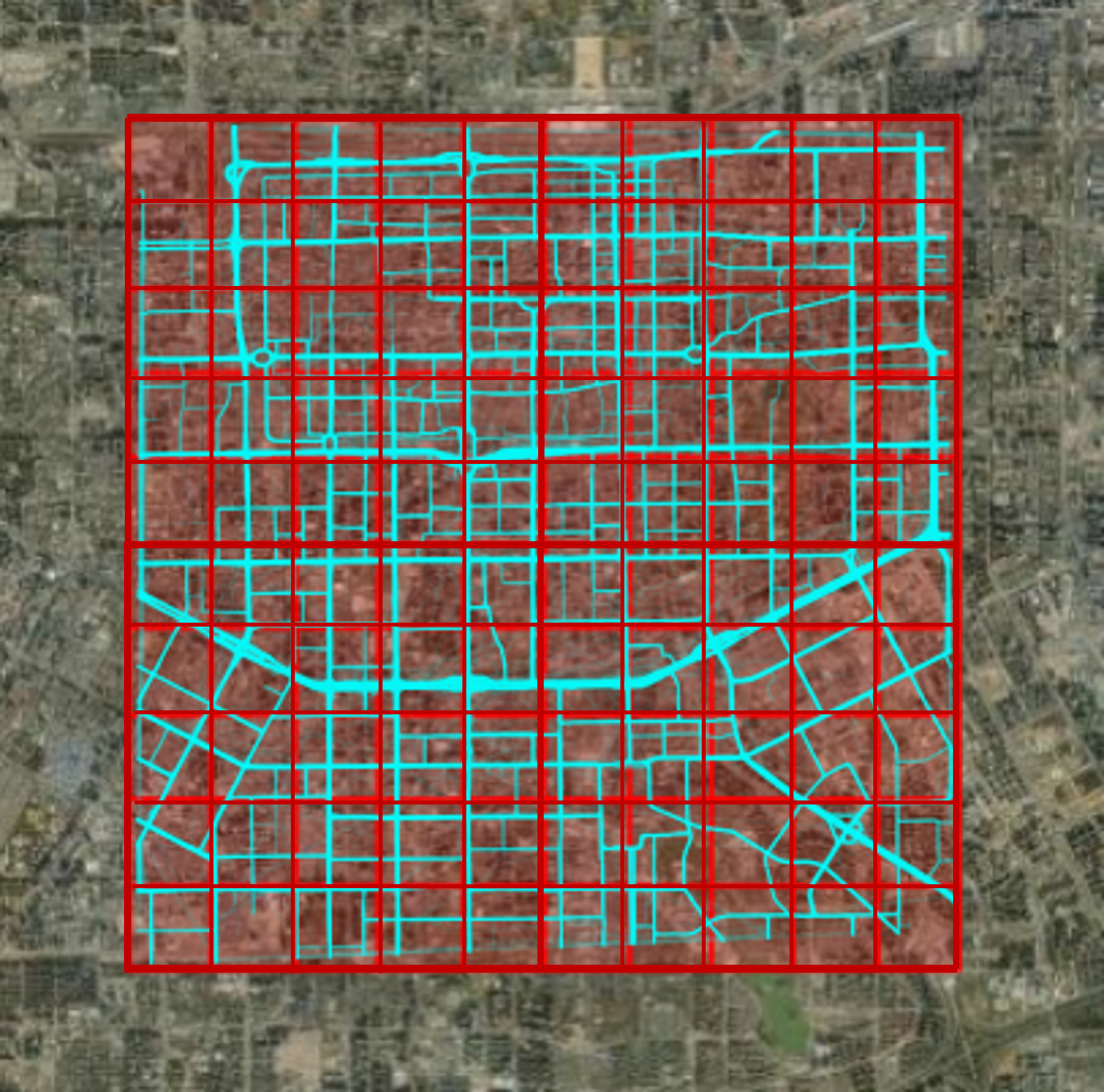} }}%
    \caption{Satellite images of Chengdu (left) and Xi'an (right), with real sample trajectories shown in cyan representing $X_c$, and where traffic flow into the red regions represents $Q_c$.}%
    \label{fig:Xian_and_Chengdu_real}%
\end{figure}

\begin{table}[h!] \footnotesize
\caption{Description of Didi Chuxing dataset for the cities of Chengdu and Xi'an.}\label{tab:Chengdu_Dataset}
\centering
\setlength\tabcolsep{2.5pt} % default value: 6pt
\begin{tabularx}{8.4cm}{ >{\centering\arraybackslash}p{2cm} | >{\centering\arraybackslash}p{3.2cm} | >{\centering\arraybackslash}p{3.2cm}}
    \toprule
    Dataset & Chengdu & Xi'an  \\
    \midrule
    Orders & 5,807,001 & 3,410,420 \\
    Drivers & 1,136,970 & 520,181 \\
    Latitudes & (30.65293, 30.72776) & (34.2053, 34.28024) \\
    Longitudes & (104.042135, 104.12959) & (108.91114, 108.9986)\\
    Time Span & \multicolumn{2}{c}{01/11/2016-30/11/2016}\\
    \bottomrule
\end{tabularx}
\end{table}

\vspace{-3mm}
\subsection{Data Pre-Processing}
The Didi Chuxing dataset is partitioned into subsets based on the number of clients emulated in the FL setting. It is assumed that each vehicle will collect data for a particular organisation, and hence, trajectories will be grouped according to the driver vehicle, and vehicles will be assigned exclusively to a particular organisation partition. For each client setting we partition vehicles evenly among the clients. Since individual drivers in a ride-sharing context will have preferences over their journeys and the number of orders they are assigned this will embed feature-based skews across the partitions and ensure realistic data distributions.

The partitioned trajectories are then processed into regional vehicle inflow format by dividing the cities into rectangular 10x10 regions, and vehicle inflows over each region are aggregated over 30 minute intervals. For the traffic flow data generation task, the partitioned traffic flow datasets, as well as the conditional dataset, are used to train TFDiff to generate synthetic traffic flow data possessing the same underlying spatio-temporal characteristics as the real data with some generational diversity. Then, FedTPS augments each client's traffic flow data with the generated synthetic data. Sequences of traffic flow data over previous 6 hours are used by the traffic prediction models to predict the regional inflow 3 hours in the future. The regional inflow data are divided into training, evaluation, and testing datasets, with the first 75\% used for training, and the final 25\% for testing.

As a measure of the heterogeneity between the regional inflow datasets, we employ three comparison metrics to assess average similarity across the partitioned client datasets: normalised Mean Absolute Difference (nMAD), Structural Similarity Index (SSI), and Temporal Correlation (TC) coefficient, where for each client and region, we calculate the temporal correlation between the two datasets over time. The properties of the partitions detailed in Table \ref{tab:Dataset_Heterogeneity_Measures} suggest that as the number of clients in a setting increases, data heterogeneity also increases; both metrics show a decrease in data similarity as the client number increases. 

\begin{table}[h!]\footnotesize
\caption{Metrics assessing the average similarity between the partitioned datasets representing each client's local dataset for each FL setting.}\label{tab:Dataset_Heterogeneity_Measures}
\centering
\setlength\tabcolsep{2pt} % default value: 6pt
\begin{tabularx}{9cm}{ >{\centering\arraybackslash}p{1cm} | >{\centering\arraybackslash}p{0.8cm} |>{\centering\arraybackslash}p{4.2cm} |>{\centering\arraybackslash}p{0.7cm}  >{\centering\arraybackslash}p{0.7cm} >{\centering\arraybackslash}p{0.7cm} }
    \toprule
    City & Clients & \# Orders per Client Partition & nMAD & SSI & TC \\
    \midrule
    \multirow{8}{*}{Chengdu} & 2 & (2904846, 2902155) & {0.1017} & {0.966} & 0.932 
    \vspace{1.6pt}\\
     & 4  & (1452404, 1452464, 1453202, 1448931) & 0.1434 & {0.936} & 0.884 \\
     & 6  & (969168, 965928, 969817, 968308, 969601, 964179) & {0.1754} & {0.908} & 0.846 \\
     & 8  & (488639, 485265, 473309, 473086, 472199, 471045, 472957, 474078) & 0.2022 & 0.883 & 0.812\\
    \midrule
    \multirow{4}{*}{Xi'an} & 2 & (1635623, 1686826) & {0.1283} & {0.916} & 0.936
    \vspace{1.6pt}\\
     & 4  & (841194, 844346, 841598, 845189) & {0.1808} & {0.851} & 0.888 \vspace{1.6pt}\\
     & 6  & (562312, 560911, 562395, 561577, 561008, 564124) & {0.2215} & {0.799} & 0.846 \\
    & 8  & (421940, 419310, 422529, 421863, 421219, 420376, 421667, 423423) & 0.2553 & 0.757 & 0.809\\
    \bottomrule
\end{tabularx}
\end{table}

  \vspace{-3mm}
\subsection{Description of Experiments}
\begin{enumerate}
    \item \textbf{Traffic Flow Data Generation:} These experiment detail the performance of the data generation models as a function of data heterogeneity and number of clients. This is important as it demonstrates the generative capability of the traffic flow generation models in federated implementations with greater data heterogeneity.
    
    \item \textbf{Traffic Flow Prediction} The purpose of this experiment is to assess how traffic flow prediction models perform over a range of different client cases under different FL frameworks, including FedTPS. This study is significant, as it emphasises how different federated mechanisms affect the performance of a global traffic flow prediction model, and highlights the benefits of a federated approach relative to traffic flow prediction models trained independently on separate data partitions.
    
    \item \textbf{Pre-Training with Synthetic Data:} This study explores the effects of pre-training the global model in an FL framework using synthetic data. This study highlights pre-training as a method for training traffic flow prediction models in scenarios with limited communication or training rounds.
    
    \item \textbf{Varying Global and Local FL Training:} The purpose of this study is to explore the benefits of FedTPS when the ratio of local training rounds to global parameter aggregation rounds is increased. Increasing the number of local training rounds per global aggregation can lead to the local training processes of each client to drift from one another, which can detrimentally impact the parameter aggregation process. 
\end{enumerate}

\subsection{Baselines}
For the traffic flow data generation task we use 300 global training rounds where we aggregate the updated client parameters after one local training round, and utilise the Adam optimiser with a learning rate of 0.0001 and a batch size of 32. We implement various baseline models including TFDiff.

\begin{enumerate}
    \item \textbf{CVAE:} A conditional VAE with a fully connected encoder and decoder, each containing three hidden layers, and featuring a latent space dimension 20.
    \item \textbf{CGAN:} A conditional GAN with three fully connected hidden layers in both the generator and discriminator, and taking a latent vector of size 100.
    \item \textbf{TFDiff:} A conditional diffusion model, with a Wide \& Deep network structure employed to embed conditional information. The encoder and decoder blocks in the UNet architecture feature multiple 1D-CNN-based stacked residual network blocks (Resnet block) with a kernel size of 3. An intermediate block, which is an attentive layer between two Resnet blocks with kernel size 1, is implemented between the encoder and decoder blocks.

\end{enumerate}

For the traffic flow prediction task we use 80 global training rounds where we aggregate the updated client parameters after one local training round, and utilise the Adam optimiser with a learning rate of 0.001 and a batch size of 32. We use various baseline models with the same model parameters as the original work, unless otherwise stated. The baselines are implemented with the following settings:

\begin{enumerate}
    \item \textbf{GRU:} A three-layer GRU model \cite{GRU} with hidden dimensions of 32 is implemented.
    \item \textbf{STGCN:} The STGCN model \cite{STGCN} using a distance based adjacency matrix.
    \item \textbf{DCRNN:} The DCRNN model \cite{DCRNN} using a distance based adjacency matrix.
    \item \textbf{GWNET:} The GWNET model \cite{GWN} using a self-adaptive adjacency matrix, where the graph connects regions within 4 kilometres.
    \item \textbf{TAU:} The TAU model \cite{TAU}.
    \item \textbf{GATAU:} The input convolutional layer is composed of two sequential 2D convolution layers with a hidden size 16. The TAU units have the same model parameters as in the original work \cite{TAU}. Our output layer is composed of two 2D convolution layers at output with hidden size 16, and a kernel size (1,1), one final linear layer to match dimensions to the output dimension. For the Graph Attention layer, the graph connects regions within 4 kilometres.
\end{enumerate}

To assess our proposed FL framework FedTPS, we implement a variety of different training frameworks. The cross-silo nature of this application motivates a setting with complete client participation; contributions from every client are critical. 

\begin{enumerate}
    \item \textbf{Non-FL Setting:} Each client trains traffic prediction model on its own partitioned dataset, and the best performing model is recorded.
    \item \textbf{FedAvg:} FedAvg \cite{FedAvg} uses a weighted averaging mechanism, based on relative client dataset sizes, to update the global model based on individual client contributions.
    \item \textbf{FedOpt:} FedOpt \cite{FedOpt} builds on FedAvg by implementing server-side adaptive optimisers such as Adam, as well as server-side momentum. For the server, we use a learning rate of 0.01, a momentum parameter of 0.9, and a second moment parameter of 0.99. 
    \item \textbf{FedProx:} FedProx \cite{FedProx} utilises a penalty term, $\mu$, to inhibit deviations of clients from the global model. We use a proximal term $\mu$ of 0.001. 
    \item \textbf{FedTPS:} The proposed framework utilises FedAvg as the parameter aggregation mechanism, and generates a synthetic dataset for the first third of the trajectory dataset, hence generates a synthetic dataset of 10 sequential days. 
\end{enumerate}

The code for this project is made available at \url{https://github.com/FerOroC/FedTPS_Traffic_Prediction}.

\section{Results}
\subsection{Traffic Flow Data Generation}
\begin{figure}[h!]
\vspace{-3mm}
  \centering
  \hspace{0.9cm}\includegraphics[width=0.85\linewidth]{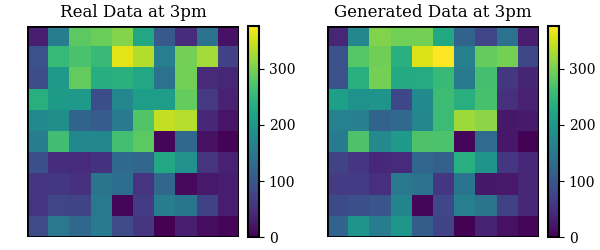}\\
  \includegraphics[width=\linewidth]{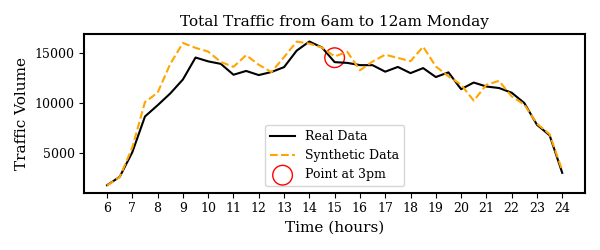}
  \caption{Heatmap visualisation of real and generated synthetic traffic flow data for the city of Chengdu from the federated TFDiff model, as well as the total traffic volumes for real and synthetic traffic flow data over all time intervals on a Monday.}
  \label{fig:Gen_data_TFDiff}
\vspace{-6mm}
\end{figure}

Figure \ref{fig:Gen_data_TFDiff} depicts heatmaps of traffic flows generated by the TFDiff compared with real traffic flow data, as well as the total region traffic flows over time intervals on a Monday. 

Table \ref{tab:TF_Model_Results} details the performance of various generative models based on three accuracy metrics which compare the similarity between the generated data, and the real data. The generative ability of the TFDiff diffusion model to generate traffic data with the same spatio-temporal characteristics as the real data is illustrated by the much lower MAD, and the very high structure similarity and temporal correlation metrics. Although the CGAN and CVAE models achieved high accuracy scores on the SSI, the high MAD and low TC scores demonstrate a high discrepancy between the real and synthetic data. This is due to the fact that cVAE and cGAN primarily capture spatial distribution, which may result in insufficient representation of the temporal dependencies inherent in traffic flow data.

\begin{table}[h!]\footnotesize
\caption{Comparison of generated regional inflow data against real regional inflow data for various centralized generative models.}
\label{tab:TF_Model_Results}
\centering
\setlength\tabcolsep{5pt} % default value: 6pt
\begin{tabularx}{6.5cm}{ >{\centering\arraybackslash}p{1.3cm} | >{\centering\arraybackslash}p{1.3cm} | >{\centering\arraybackslash}p{0.85cm} >{\centering\arraybackslash}p{0.85cm} >{\centering\arraybackslash}p{0.55cm}   >{\centering\arraybackslash}p{1.3cm}}
    \toprule  
    City & Model & MAD & SSI & TC  \\
    \toprule
    \multirow{3}{*}{Chengdu} 
    & CGAN & 0.53 & 0.86 & 0.33 \\
    & CVAE & 0.55 & 0.93 & 0.53 \\
    & TFDiff & 0.14 & 0.97 & 0.92 \\
    \midrule
    \multirow{3}{*}{Xi'an} 
    & CGAN  & 1.02 & 0.78 & 0.22 \\
    & CVAE  & 1.40 & 0.88 & 0.29 \\
    & TFDiff  & 0.26 & 0.93 & 0.91 \\
    \bottomrule
\end{tabularx}
\vspace{-2mm}
\end{table}

For the TFDiff model trained under federated settings with a varying number of clients, Table \ref{tab:TF_Model_FL_Results} shows the similarity between generated samples and real traffic data based on four metrics, averaged over the regional inflow data sequences. Some deviation is expected due to the generational diversity of the DiffTraj model, yet the high SSI and TC metrics demonstrate that the spatio-temporal characteristics of the original data were modelled accurately by the trajectory generation model. In general, increasing the number of clients in the FL setting results in a larger deviation from the real traffic data for the samples generated, as evidenced by the decreasing values of SSI and Temporal Correlation, as well as increasing nMAE and MAPE metrics.

\begin{table}[h!]\footnotesize
\caption{Comparison of generated regional inflow data against real regional inflow data for the TFDiff model trained through FL, with varying client cases.}
\label{tab:TF_Model_FL_Results}
\centering
\setlength\tabcolsep{5pt} % default value: 6pt
\begin{tabularx}{7.5cm}{ >{\centering\arraybackslash}p{1.3cm} | >{\centering\arraybackslash}p{1.3cm} | >{\centering\arraybackslash}p{0.85cm} >{\centering\arraybackslash}p{0.85cm} >{\centering\arraybackslash}p{0.55cm}   >{\centering\arraybackslash}p{0.55cm}}
    \toprule  
    City & Clients & nMAE & MAPE & SSI & TC  \\
    \toprule
    \multirow{4}{*}{Chengdu} & 2  & 0.13 & 16.67 & 0.97 & 0.91 \\
    & 4  & 0.18 & 21.21 & 0.93 & 0.86 \\
    & 6  & 0.20  & 22.26 & 0.91 & 0.82 \\
    & 8  & 0.23  & 23.77 & 0.88 & 0.79 \\
    \midrule
    \multirow{4}{*}{Xi'an} & 2  & 0.31 & 31.25 & 0.94 &  0.88\\
    & 4  & 0.33 & 29.04 & 0.91 & 0.85 \\
    & 6  & 0.38 & 29.90 & 0.88 & 0.81 \\
    & 8  & 0.43 & 31.47 & 0.86 & 0.78 \\
    \bottomrule
\end{tabularx}
\end{table}

\begin{table*}[tp]\centering \footnotesize
    \caption{Comparison of various Spatio-Temporal model performances for prediction of traffic at 6 time-steps in future. Assessed using nMAE and MAPE metrics, under varying client numbers and FL methods. (All numeric values are shown to 4 significant figures.)}
    \label{Chengdu_Table}
    \setlength\tabcolsep{5.5pt} % Reduce column separation
    \begin{tabular}{ >{\centering\arraybackslash}p{1.1cm} |
                     >{\centering\arraybackslash}p{0.9cm} |
                     >{\centering\arraybackslash}p{1.5cm} |
                     >{\centering\arraybackslash}p{0.55cm}
                     >{\centering\arraybackslash}p{0.85cm} |
                     >{\centering\arraybackslash}p{0.55cm}
                     >{\centering\arraybackslash}p{0.85cm} |
                     >{\centering\arraybackslash}p{0.55cm}
                     >{\centering\arraybackslash}p{0.85cm} |
                     >{\centering\arraybackslash}p{0.65cm}
                     >{\centering\arraybackslash}p{0.85cm} |
                     >{\centering\arraybackslash}p{0.55cm}
                     >{\centering\arraybackslash}p{0.85cm} |
                     >{\centering\arraybackslash}p{0.55cm}
                     >{\centering\arraybackslash}p{0.85cm}}
    \toprule
    & & & \multicolumn{2}{c|}{GRU}
        & \multicolumn{2}{c|}{STGCN}
        & \multicolumn{2}{c|}{DCRNN}
        & \multicolumn{2}{c|}{GWNET}
        & \multicolumn{2}{c|}{TAU}
        & \multicolumn{2}{c}{GATAU} \\
    City & Clients & FL Method 
          & nMAE & MAPE 
          & nMAE & MAPE 
          & nMAE & MAPE 
          & nMAE & MAPE 
          & nMAE & MAPE 
          & nMAE & MAPE \\
    \toprule
    %-------------------- Chengdu ----------------------
\multirow{20}{*}{Chengdu}
& \multirow{5}{*}{2}
  & Non-FL
    & 0.2385 & 35.54 
    & 0.1601 & 32.98 
    & 0.1528 & 27.56 
    & 0.1437 & 24.57 
    & 0.1145 & 20.66 
    & 0.1205 & 22.74 \\
  \cdashline{3-15}
  & & FedAvg
    & 0.2187 & 24.03 
    & 0.1249 & 23.21 
    & 0.1174 & 20.56 
    & 0.1145 & 19.86 
    & 0.1054 & 17.85 
    & 0.1142 & 19.29 \\
  & & FedOpt
    & 0.2190 & 23.90 
    & 0.1186 & 21.88 
    & 0.1152 & 20.26 
    & 0.1149 & 20.19 
    & 0.1043 & 17.91 
    & 0.1144 & 19.01 \\
  & & FedProx
    & 0.2167 & 25.02 
    & 0.1215 & 23.07 
    & 0.1214 & 21.10 
    & 0.1186 & 21.28 
    & 0.1122 & 18.61 
    & 0.1129 & 18.88 \\
  & & FedTPS
    & \textbf{0.2143} & \textbf{22.91}
    & \textbf{0.1122} & \textbf{19.88}
    & \textbf{0.1065} & \textbf{17.96}
    & \textbf{0.1080} & \textbf{18.69}
    & \textbf{0.1019} & \textbf{17.11}
    & \textbf{0.1090} & \textbf{18.02} \\
\cline{2-15}
& \multirow{5}{*}{4}
  & Non-FL
    & 0.2512 & 27.65 
    & 0.1758 & 30.47 
    & 0.1750 & 26.91 
    & 0.1651 & 25.01 
    & 0.1487 & 22.66 
    & 0.1472 & 21.59 \\
  \cdashline{3-15}
  & & FedAvg
    & 0.2331 & 24.31 
    & 0.1426 & 23.72 
    & 0.1394 & 22.67 
    & 0.1362 & 21.73 
    & \textbf{0.1285} & 20.62 
    & 0.1369 & 21.42 \\
  & & FedOpt
    & 0.2329 & 24.33 
    & 0.1442 & 24.26 
    & 0.1426 & 22.94 
    & 0.1384 & 22.02 
    & 0.1289 & \textbf{20.44}
    & 0.1373 & 21.22 \\
  & & FedProx
    & 0.2325 & 24.72 
    & 0.1455 & 24.75 
    & 0.1430 & 22.93 
    & 0.1386 & 22.38 
    & 0.1341 & 21.36 
    & 0.1373 & 21.21 \\
  & & FedTPS
    & \textbf{0.2272} & \textbf{23.28}
    & \textbf{0.1360} & \textbf{22.62}
    & \textbf{0.1316} & \textbf{21.22}
    & \textbf{0.1282} & \textbf{20.39}
    & 0.1335 & 20.89
    & \textbf{0.1299} & \textbf{20.38} \\
\cline{2-15}
& \multirow{5}{*}{6}
  & Non-FL
    & 0.2614 & 32.31 
    & 0.1962 & 33.88 
    & 0.1954 & 31.03 
    & 0.1877 & 28.70 
    & 0.1760 & 26.66 
    & 0.1660 & 26.01 \\
  \cdashline{3-15}
  & & FedAvg
    & 0.2462 & 25.54 
    & 0.1588 & 24.58 
    & 0.1602 & 23.59 
    & 0.1557 & 23.02 
    & \textbf{0.1485} & \textbf{21.86}
    & 0.1571 & 22.78 \\
  & & FedOpt
    & 0.2459 & 25.35 
    & 0.1602 & 24.99 
    & 0.1614 & 24.01 
    & 0.1554 & 23.02 
    & 0.1488 & 21.91 
    & 0.1562 & 22.48 \\
  & & FedProx
    & 0.2442 & 25.65 
    & 0.1614 & 25.45 
    & 0.1620 & 24.23 
    & 0.1557 & 23.39 
    & 0.1528 & 22.57 
    & 0.1557 & 22.66 \\
  & & FedTPS
    & \textbf{0.2402} & \textbf{24.57}
    & \textbf{0.1551} & \textbf{24.09}
    & \textbf{0.1525} & \textbf{22.62}
    & \textbf{0.1502} & \textbf{22.15}
    & 0.1542 & 22.69
    & \textbf{0.1502} & \textbf{22.05} \\
\cline{2-15}
& \multirow{5}{*}{8}
  & Non-FL
    & 0.2738 & 28.08 
    & 0.2091 & 31.36 
    & 0.2114 & 29.24 
    & 0.2057 & 28.62 
    & 0.2038 & 27.52 
    & 0.1836 & 25.78 \\
  \cdashline{3-15}
  & & FedAvg
    & 0.2590 & 27.17 
    & 0.1813 & 26.73 
    & 0.1824 & 26.02 
    & 0.1748 & 24.98 
    & 0.1691 & 24.00 
    & 0.1771 & 24.45 \\
  & & FedOpt
    & 0.2586 & 27.23 
    & 0.1805 & 27.14 
    & 0.1813 & 25.62 
    & 0.1756 & 25.64 
    & \textbf{0.1687} & \textbf{23.79}
    & 0.1760 & 24.47 \\
  & & FedProx
    & 0.2597 & 27.18 
    & 0.1855 & 27.81 
    & 0.1832 & 25.99 
    & 0.1763 & 25.34 
    & 0.1706 & 24.23 
    & 0.1782 & 24.74 \\
  & & FedTPS
    & \textbf{0.2556} & \textbf{26.83}
    & \textbf{0.1767} & \textbf{26.50}
    & \textbf{0.1741} & \textbf{24.60}
    & \textbf{0.1691} & \textbf{24.39}
    & 0.1801 & 24.87
    & \textbf{0.1725} & \textbf{23.79} \\
\cline{1-15} \cline{2-15}
%-------------------- Xi'an -----------------------------
\multirow{20}{*}{Xi'an}
& \multirow{5}{*}{2}
  & Non-FL
    & 0.2389 & 44.62 
    & 0.1719 & 30.44 
    & 0.1716 & 28.74 
    & 0.1516 & 24.91 
    & 0.1387 & 23.25 
    & 0.1350 & 21.21 \\
  \cdashline{3-15}
  & & FedAvg
    & 0.2127 & 33.05 
    & 0.1406 & 20.57 
    & 0.1344 & 20.73 
    & 0.1357 & 21.24 
    & \textbf{0.1271} & 19.08 
    & 0.1298 & 19.88 \\
  & & FedOpt
    & 0.2133 & 33.44 
    & 0.1369 & 20.13 
    & 0.1351 & 20.40 
    & 0.1395 & 21.51 
    & 0.1276 & 18.81 
    & 0.1276 & 20.18 \\
  & & FedProx
    & 0.2151 & 33.31 
    & 0.1399 & 21.56 
    & 0.1352 & 21.29 
    & 0.1376 & 21.63 
    & 0.1324 & 20.35 
    & 0.1300 & 21.05 \\
  & & FedTPS
    & \textbf{0.2096} & \textbf{32.09}
    & \textbf{0.1256} & \textbf{18.23}
    & \textbf{0.1284} & \textbf{19.59}
    & \textbf{0.1302} & \textbf{19.96}
    & 0.1304 & \textbf{18.60}
    & \textbf{0.1240} & \textbf{19.35} \\
\cline{2-15}
& \multirow{5}{*}{4}
  & Non-FL
    & 0.2587 & 38.18 
    & 0.2173 & 34.52 
    & 0.2071 & 32.34 
    & 0.1871 & 28.58 
    & 0.1852 & 26.99 
    & 0.1657 & 25.11 \\
  \cdashline{3-15}
  & & FedAvg
    & 0.2348 & 27.00 
    & 0.1706 & 23.08 
    & 0.1682 & 23.58 
    & 0.1632 & 23.38 
    & \textbf{0.1618} & \textbf{22.15}
    & 0.1605 & 22.34 \\
  & & FedOpt
    & 0.2354 & 27.47 
    & 0.1698 & 22.95 
    & 0.1682 & 23.53 
    & 0.1632 & 23.41 
    & 0.1627 & 22.32 
    & 0.1597 & 22.40 \\
  & & FedProx
    & 0.2362 & 27.17 
    & 0.1736 & 23.93 
    & 0.1693 & 23.39 
    & 0.1632 & 23.53 
    & \textbf{0.1618} & 22.95
    & 0.1594 & 22.43 \\
  & & FedTPS
    & \textbf{0.2345} & \textbf{26.68}
    & \textbf{0.1621} & \textbf{22.06}
    & \textbf{0.1613} & \textbf{22.94}
    & \textbf{0.1569} & \textbf{22.64}
    & 0.1709 & 23.37
    & \textbf{0.1586} & \textbf{21.81} \\
\cline{2-15}
& \multirow{5}{*}{6}
  & Non-FL
    & 0.2819 & 33.64 
    & 0.2230 & 32.15 
    & 0.2271 & 32.56 
    & 0.2099 & 30.39 
    & 0.2271 & 31.77 
    & 0.1913 & 26.23 \\
  \cdashline{3-15}
  & & FedAvg
    & 0.2535 & 26.35 
    & 0.1963 & 26.02 
    & 0.1975 & 25.92 
    & 0.1930 & 26.35 
    & \textbf{0.1856} & 24.40
    & 0.1823 & \textbf{24.43} \\
  & & FedOpt
    & 0.2535 & 26.52 
    & 0.1938 & 25.96 
    & 0.1975 & 25.69 
    & 0.1930 & 26.82 
    & 0.1872 & \textbf{24.25}
    & \textbf{0.1819} & 24.54 \\
  & & FedProx
    & 0.2559 & 26.69 
    & 0.1983 & 26.85 
    & 0.1983 & 26.08 
    & 0.1959 & 26.49 
    & 0.1880 & 25.05 
    & 0.1839 & 24.83 \\
  & & FedTPS
    & \textbf{0.2510} & \textbf{26.26}
    & \textbf{0.1893} & \textbf{25.79}
    & \textbf{0.1913} & \textbf{25.39}
    & \textbf{0.1913} & 26.34
    & 0.2020 & 25.91
    & 0.1827 & 24.60 \\
\cline{2-15}
& \multirow{5}{*}{8}
  & Non-FL
    & 0.2935 & 28.92 
    & 0.2403 & 31.80 
    & 0.2480 & 30.99 
    & 0.2359 & 30.10 
    & 0.2529 & 31.66 
    & 0.2238 & 27.41 \\
  \cdashline{3-15}
  & & FedAvg
    & 0.2688 & 27.76
    & 0.2129 & 28.26
    & 0.2173 & 28.30
    & 0.2107 & 27.99
    & 0.2107 & \textbf{26.39}
    & 0.2063 & \textbf{26.30} \\
  & & FedOpt
    & 0.2694 & 27.79
    & 0.2129 & 28.20
    & 0.2173 & 28.01
    & 0.2085 & 27.92
    & \textbf{0.2090} & 26.52
    & \textbf{0.2052} & 26.70 \\
  & & FedProx
    & 0.2683 & 28.02
    & 0.2140 & 28.63
    & 0.2184 & 28.40
    & 0.2101 & 27.94
    & 0.2096 & 26.99
    & 0.2057 & 26.71 \\
  & & FedTPS
    & \textbf{0.2677} & \textbf{27.74}
    & \textbf{0.2107} & \textbf{28.10}
    & \textbf{0.2129} & \textbf{27.68}
    & \textbf{0.2074} & \textbf{27.70}
    & 0.2255 & 28.02
    & 0.2057 & 26.31 \\
    \bottomrule
    \end{tabular}
\vspace{-5mm}
\end{table*}

Table \ref{tab:Dataset_Heterogeneity_Measures} shows that the data heterogeneity in the partitioned datasets increases when there are more clients in the federated setting. This can cause the global model to struggle generalising across the multiple data silos, which leads to model performance degradation with higher client numbers, and slower global model convergence. Yet the samples generated throughout all client cases demonstrate high degrees of similarity with the real data, as the structural similarity index and temporal correlation among regions remain close to 1, and the nRMSE remains low throughout the different cases.

\subsection{Traffic Flow Prediction}
Table \ref{Chengdu_Table} presents the results of the traffic prediction models, trained under different FL paradigms with varying client numbers, with respect to nMAE and MAPE metrics. FedTPS consistently outperforms other FL frameworks for the Chengdu dataset. Our reasoning for FedTPS's improved performance relative to other FL frameworks is due to the increased amount of data available to the clients during training, as well as the data heterogeneity reduction stemming from the shared synthetic dataset. Through synthetic data augmentation, we ensure that each client's dataset is more representative of the overall data distribution, and increase the amount of information available to each client.

Comparing FedAvg and FedOpt, there is no noticeable difference between the two mechanisms. This is consistent with wider research on cross-silo FL with real-world data, which found that no State-of-the-Art FL framework consistently outperformed FedAvg \cite{Results_SOTA_FL}. FedProx generally performs worse than the other FL frameworks. It should be noted that the data augmentation framework of FedTPS can be used in conjunction to other FL framework which optimise the aggregation protocol for global model updates or client selection strategy, such as any of the FL baselines presented in this work. 

With regards to the traffic flow prediction models, TAU consistently outperforms other State-of-the-Art models. When trained with FedTPS, GATAU can yield better results, as shown in all the client cases for the Xi'an dataset. The Graph Attention mechanism can learn cross-region relationships from the entire city, whereas the convolutional operations implemented in the TAU restrict spatial dependencies modelling to neighbouring regions. Therefore, data augmentation may benefit GATAU more, as there is more data available to learn cross-region dependencies.

Aside from the two client case for both cities, training the TAU model under the FedTPS framework degrades the performance of the global model relative to FedAvg. This may stem from the fact that the TAU model mainly focuses on capturing temporal dependencies of the data, and Table \ref{tab:TF_Model_FL_Results} shows that TC decreases more significantly than SSI in higher client cases. 

At lower client numbers, the benefits of FedTPS are more noticeable. This is likely due to the performance of the trajectory generation model improving at low client numbers, hence leading to higher quality synthetic trajectories generated. Furthermore, as shown in Table \ref{tab:Dataset_Heterogeneity_Measures}, data heterogeneity increases when partitioning the data for higher client number cases, and thus, the optimisation objective in local training for a client varies more. This impacts the federated global model, which may result in worse performing global models. 

\vspace{-3mm}
\subsection{Communication Costs}
Table \ref{tab:Communication_Cost} presents the communication costs for the distribution of the global model to each client in the federated setting, for TFDiff and the traffic prediction models. The generative TFDiff model which is trained in the first stage of FedTPS requires less communication cost than TAU or GATAU. Table \ref{tab:Communication_Cost} also demonstrates that the GATAU model incurs less communication costs than the TAU model. It should be noted that for cross-silo FL settings, communication bottlenecks are not as prevalent; server-to-client connections represent high-bandwidth data center network connections. Furthermore, complete client participation is important for cross-silo settings as contributions from every client are critical.

\begin{table}[H]\centering \small
\caption{Communication costs of FedAvg and FedTPS for selected traffic flow prediction models for different client numbers, in MB.}
\label{tab:Communication_Cost}
\setlength\tabcolsep{6pt} % default value: 6pt
\begin{tabular}{ >{\centering\arraybackslash}p{0.85cm} | >{\centering\arraybackslash}p{0.85cm} >{\centering\arraybackslash}p{0.85cm}  >{\centering\arraybackslash}p{0.85cm}  >{\centering\arraybackslash}p{0.85cm}>{\centering\arraybackslash}p{0.85cm}  >{\centering\arraybackslash}p{0.85cm} }
    \toprule  
   \multicolumn{7}{c}{Comm. Cost per Client per FL Round} \\
     TFDiff & GRU & STGCN & DCRNN & GWNET & TAU & GATAU  \\
    \midrule
     {47.52} & {0.27} & {0.87} & {1.79} & {1.24} & 50.76 & 48.40 \\
    \bottomrule
\end{tabular}
\end{table}

\vspace{-5mm}
\subsection{Effects of Synthetic Pre-training on Global Model Training}
\begin{figure}[htbp]\small
    \centering
    \includegraphics[width=8cm]{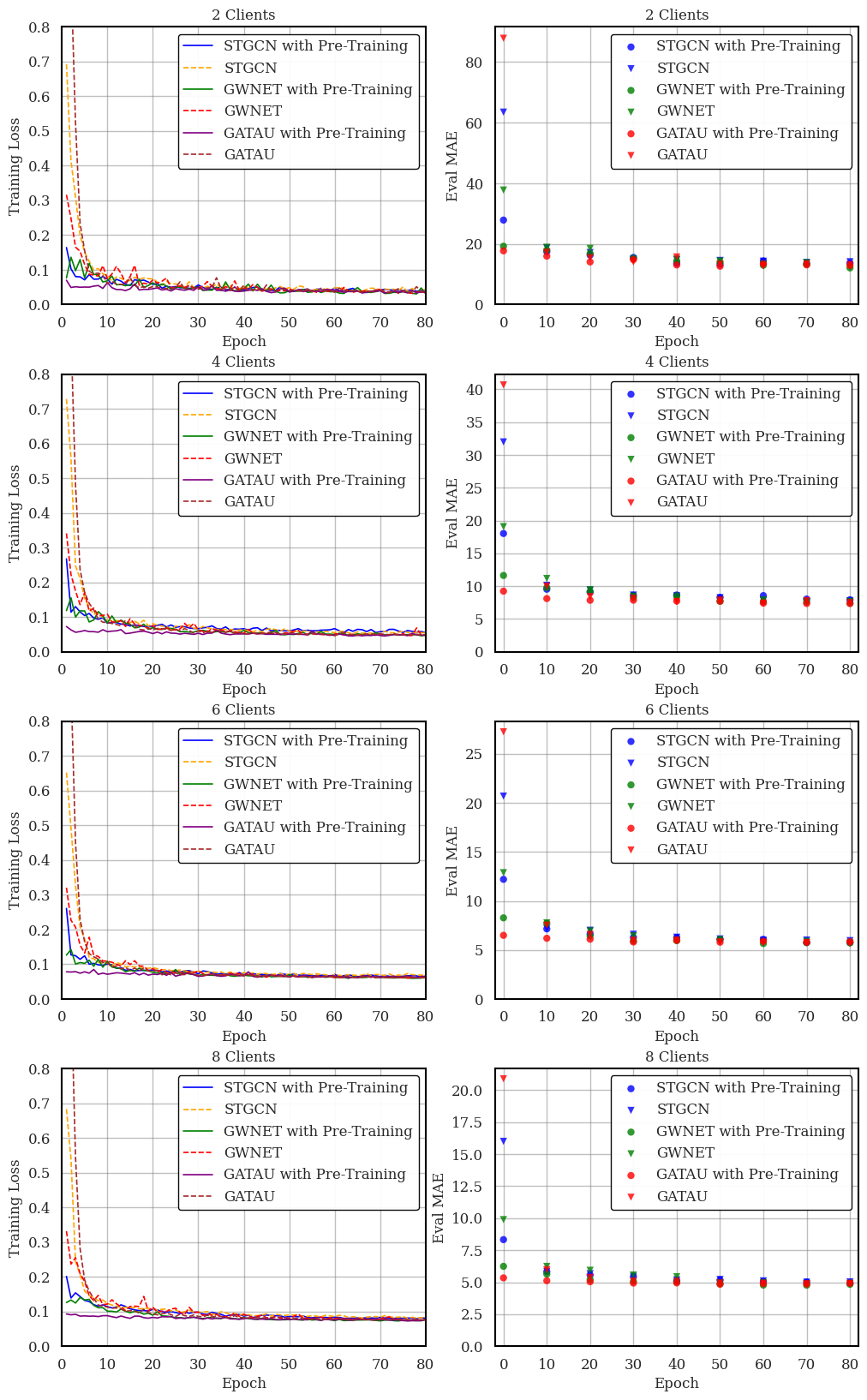}
    \caption{Effects of Pre-Training models with Synthetic Data on model training process.}
    \label{fig:Pre_training_effects}
\vspace{-5mm}
\end{figure}

Figure \ref{fig:Pre_training_effects} details the training loss and evaluation metrics of select traffic prediction models under FL settings, with and without pre-training. With pre-training, the models converge more rapidly than the FL settings without pre-training. However, by global round 60, the models will have generally converged to the same training loss. Hence, while pre-training with synthetic data improves the convergence speed of the global model noticeably, it offers no improved model performance. The random initialisation of model parameters causes the training loss and evaluation metrics at global round 1 to be very high for models without pre-training. However, with pre-training, the starting point of the model is competitive, as can be seen from the low metric scores at global round 1.

\subsection{Effect of Client Rounds and Server Rounds}
\begin{table}[h!]\centering \footnotesize
    \caption{Comparison of selected traffic flow prediction models trained in a federated setting employing FedAvg and FedTPS as the ratio of local and global training rounds are varied.}
    \label{Local_vs_Global_Rounds}
    \setlength\tabcolsep{4pt} % default value: 6pt
    \begin{tabular}{ >{\centering\arraybackslash}p{0.8cm} | >{\centering\arraybackslash}p{1cm} | >{\centering\arraybackslash}p{1cm} | >{\centering\arraybackslash}p{1.4cm} | >{\centering\arraybackslash}p{0.9cm} >{\centering\arraybackslash}p{0.9cm}}
    \toprule
    Clients & Model & FL Method & $r_{loc}:r_{\textnormal{glob}}$ & nMAE & MAPE  \\
    \midrule
    %% 2 clients
\multirow{12}{*}{2} & \multirow{4}{*}{STGCN} & \multirow{2}{*}{FedAvg} & 10 & 0.1213 & 23.10 \\
 &  &  & 20 & 0.1210 & 23.23 \\
\cline{3-6}
 &  & \multirow{2}{*}{FedTPS} & 10 & 0.1153 & 20.12 \\
 &  &  & 20 & 0.1145 & 20.58 \\
\cline{2-6} \cline{3-6}
 & \multirow{4}{*}{GWNET} & \multirow{2}{*}{FedAvg} & 10 & 0.1113 & 18.88 \\
 &  &  & 20 & 0.1163 & 19.70 \\
\cline{3-6}
 &  & \multirow{2}{*}{FedTPS} & 10 & 0.1085 & 17.95 \\
 &  &  & 20 & 0.1097 & 19.22 \\
\cline{2-6} \cline{3-6}
 & \multirow{4}{*}{GATAU} & \multirow{2}{*}{FedAvg} & 10 & 0.1192 & 20.46 \\
 &  &  & 20 & 0.1124 & 20.08 \\
\cline{3-6}
 &  & \multirow{2}{*}{FedTPS} & 10 & 0.1101 & 17.93 \\
 &  &  & 20 & 0.1107 & 18.21 \\
\cline{1-6} \cline{2-6} \cline{3-6}
\multirow{12}{*}{4} & \multirow{4}{*}{STGCN} & \multirow{2}{*}{FedAvg} & 10 & 0.1466 & 23.76 \\
 &  &  & 20 & 0.1480 & 24.58 \\
\cline{3-6}
 &  & \multirow{2}{*}{FedTPS} & 10 & 0.1410 & 23.68 \\
 &  &  & 20 & 0.1414 & 23.39 \\
\cline{2-6} \cline{3-6}
 & \multirow{4}{*}{GWNET} & \multirow{2}{*}{FedAvg} & 10 & 0.1384 & 21.85 \\
 &  &  & 20 & 0.1383 & 21.79 \\
\cline{3-6}
 &  & \multirow{2}{*}{FedTPS} & 10 & 0.1277 & 21.20 \\
 &  &  & 20 & 0.1308 & 21.23 \\
\cline{2-6} \cline{3-6}
 & \multirow{4}{*}{GATAU} & \multirow{2}{*}{FedAvg} & 10 & 0.1374 & 20.98 \\
 &  &  & 20 & 0.1358 & 21.17 \\
\cline{3-6}
 &  & \multirow[t]{2}{*}{FedTPS} & 10 & 0.1322 & 20.98 \\
 &  &  & 20 & 0.1329 & 20.39 \\
\cline{1-6} \cline{2-6} \cline{3-6}
\multirow{12}{*}{6} & \multirow{4}{*}{STGCN} & \multirow{2}{*}{FedAvg} & 10 & 0.1633 & 25.95 \\
 &  &  & 20 & 0.1637 & 25.76 \\
\cline{3-6}
 &  & \multirow{2}{*}{FedTPS} & 10 & 0.1572 & 24.45 \\
 &  &  & 20 & 0.1579 & 24.53 \\
\cline{2-6} \cline{3-6}
 & \multirow{4}{*}{GWNET} & \multirow{2}{*}{FedAvg} & 10 & 0.1563 & 23.04 \\
 &  &  & 20 & 0.1582 & 23.58 \\
\cline{3-6}
 &  & \multirow{2}{*}{FedTPS} & 10 & 0.1502 & 22.26 \\
 &  &  & 20 & 0.1525 & 22.24 \\
\cline{2-6} \cline{3-6}
 & \multirow{4}{*}{GATAU} & \multirow{2}{*}{FedAvg} & 10 & 0.1559 & 22.33 \\
 &  &  & 20 & 0.1562 & 22.51 \\
\cline{3-6}
 &  & \multirow{2}{*}{FedTPS} & 10 & 0.1516 & 22.04 \\
 &  &  & 20 & 0.1499 & 21.80 \\
\cline{1-6} \cline{2-6} \cline{3-6}
\multirow{12}{*}{8} & \multirow{4}{*}{STGCN} & \multirow{2}{*}{FedAvg} & 10 & 0.1860 & 28.16 \\
 &  &  & 20 & 0.1843 & 27.89 \\
\cline{3-6}
 &  & \multirow{2}{*}{FedTPS} & 10 & 0.1801 & 26.91 \\
 &  &  & 20 & 0.1854 & 27.73 \\
\cline{2-6} \cline{3-6}
 & \multirow{4}{*}{GWNET} & \multirow{2}{*}{FedAvg} & 10 & 0.1764 & 25.01 \\
 &  &  & 20 & 0.1791 & 25.52 \\
\cline{3-6}
 &  & \multirow{2}{*}{FedTPS} & 10 & 0.1710 & 24.57 \\
 &  &  & 20 & 0.1708 & 24.43 \\
\cline{2-6} \cline{3-6}
 & \multirow{4}{*}{GATAU} & \multirow{2}{*}{FedAvg} & 10 & 0.1772 & 24.50 \\
 &  &  & 20 & 0.1775 & 24.14 \\
\cline{3-6}
 &  & \multirow{2}{*}{FedTPS} & 10 & 0.1715 & 23.96 \\
 &  &  & 20 & 0.1734 & 23.75 \\
     \cline{1-6}
    \end{tabular}
\vspace{-3mm}
\end{table}

Table \ref{Local_vs_Global_Rounds} illustrates the results of this study, which explores varying the number of local training rounds per global aggregation step. FedTPS consistently demonstrates lowest error metrics than FedAvg. The decrease in data heterogeneity stemming from the shared synthetic dataset, coupled with the increased information available to each client, means that even in cases where there are greater local training rounds relative to global aggregation rounds, the global model performance is enhanced. This may be because the synthetic data reduces the variance between client updates and promotes more stable and consistent parameter updates. Thus, FedTPS improves the robustness of the FL process against the challenges posed by non-IID data distributions and limited communication rounds.

\section{Conclusion}

This paper addresses the task of Federated Learning for traffic flow prediction applications, and presents a new framework referred to as FedTPS which facilitates synthetic data augmentation of client datasets. By reducing data heterogeneity among clients and enhancing the information available during local training, FedTPS is able to consistently outperform other FL mechanisms for a variety of different client settings and different traffic flow prediction models. An FL setting to traffic flow prediction tasks is emulated using a large-scale ride sharing dataset for the cities of Chengdu and Xi'an. The trajectory data is partitioned into subsets to emulate individual organisations which collect their own data. Organisations can then collaborate to train models in a federated manner, without infringing upon data privacy. FedTPS first trains a federated synthetic data generation model, TFDiff, and then generates a synthetic regional traffic flow dataset from the learned global distribution of trajectories. This synthetic data is then distributed to all clients prior to the traffic flow prediction task so they may augment their local datasets. Future work could apply FedTPS in highly non-IID data settings.

\bibliographystyle{IEEEtran}
\bibliography{main}

 \begin{IEEEbiography}
 [{\includegraphics[width=1in,height=1.25in,clip,keepaspectratio]{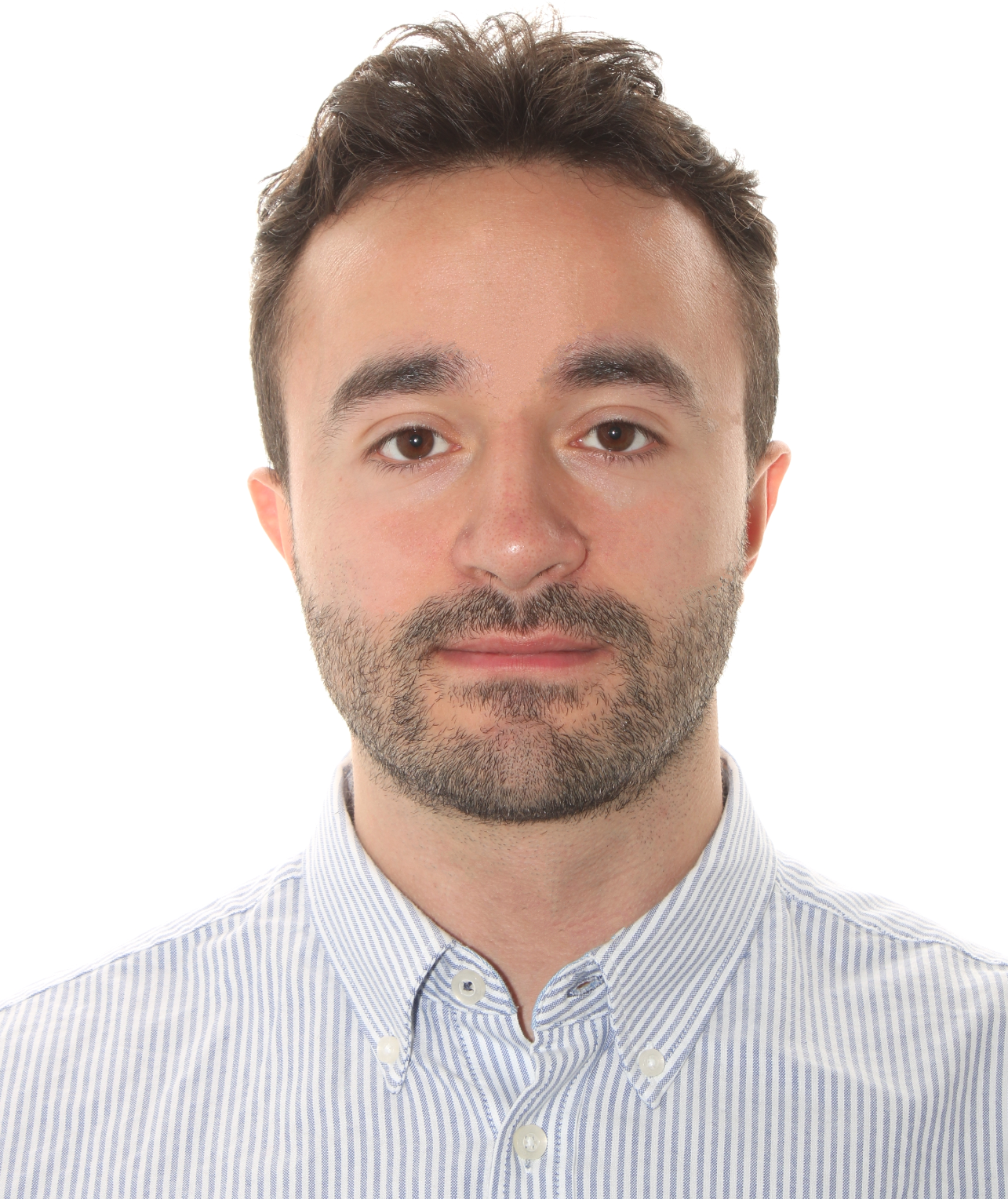}}]{Fermin Orozco} is currently a PhD student within the Department of Computer Science at the University of Exeter. He received the MEng degree from the University of Bristol in 2021, and prior to his PhD, worked within the Intelligent Transport Systems team at AECOM. His research interests include Federated Learning applications within Intelligent Transport Systems.
\end{IEEEbiography}

\begin{IEEEbiography}[{\includegraphics[width=1in,height=1.25in,clip,keepaspectratio]{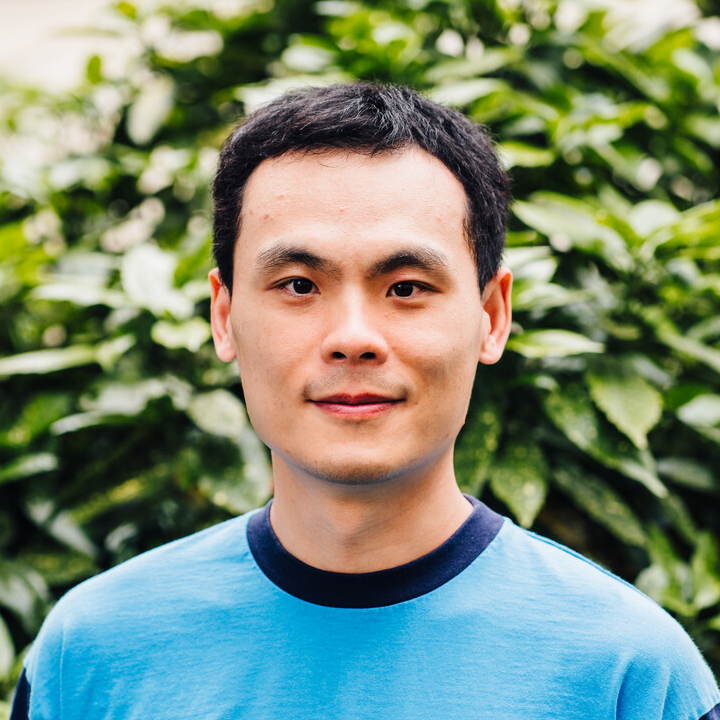}}]{Hongkai Wen} received the D.Phil. degree from the University of Oxford, where he was a Postdoctoral Researcher with the Oxford Computer Science and Robotics Institute. He is currently a Professor with the Department of Computer Science, University of Warwick. His research interests include mobile sensor systems, human-centric sensing, and pervasive data science.

\end{IEEEbiography}

 \begin{IEEEbiography}
 [{\includegraphics[width=1in,height=1.25in,clip,keepaspectratio]{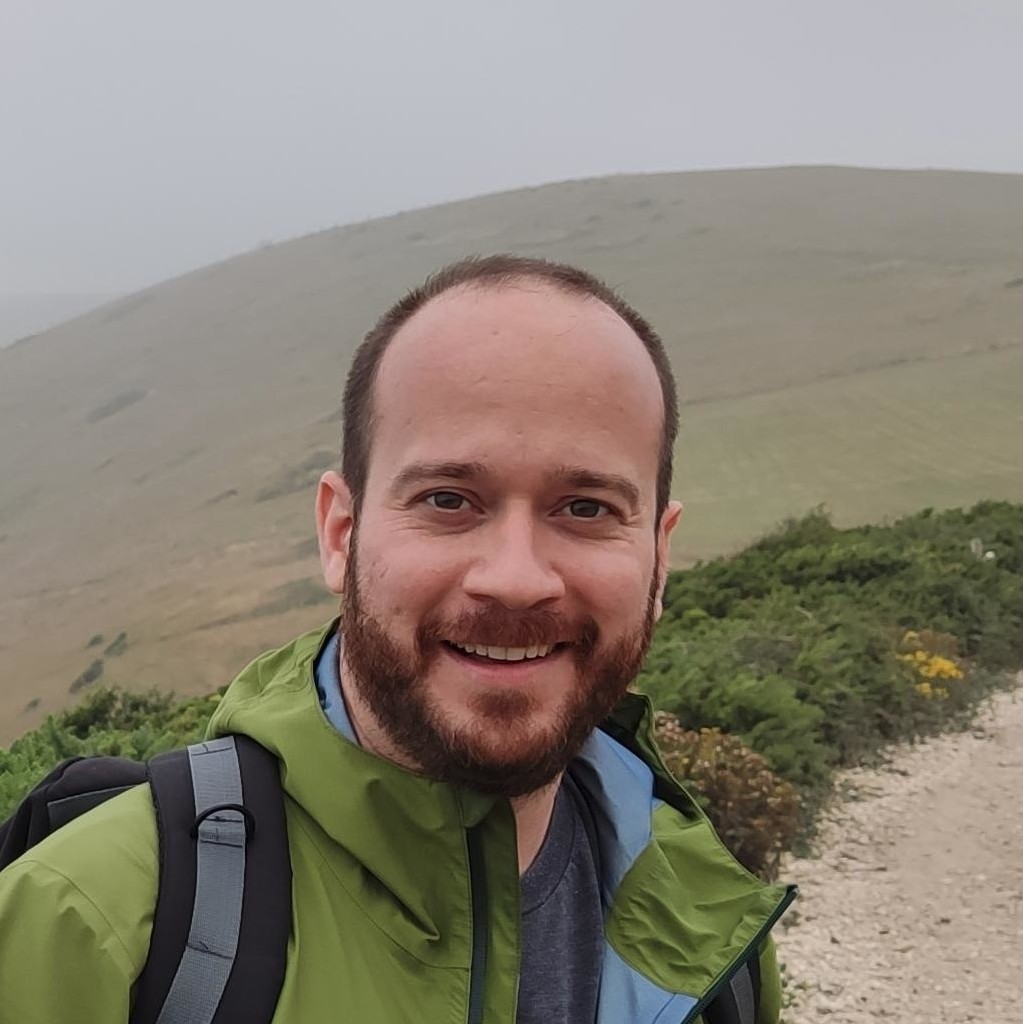}}]
 {Pedro Porto Buarque de Gusmão} received the B.Sc. degree from the University of São Paulo, M.Sc. degree from the Polytechnic University of Turin, and Ph.D. degree from the Polytechnic University of Turin, in 2017. He is currently a Lecturer in Computer Science at the University of Surrey. He has previously held Postdoctoral Researcher roles with the University of Oxford, and the University of Cambridge. His research interests include Computer Vision, Navigation under adverse conditions, Sensor Fusion, and Distributed Machine Learning.
\end{IEEEbiography}

 \begin{IEEEbiography}
 [{\includegraphics[width=1in,height=1.25in,clip,keepaspectratio]{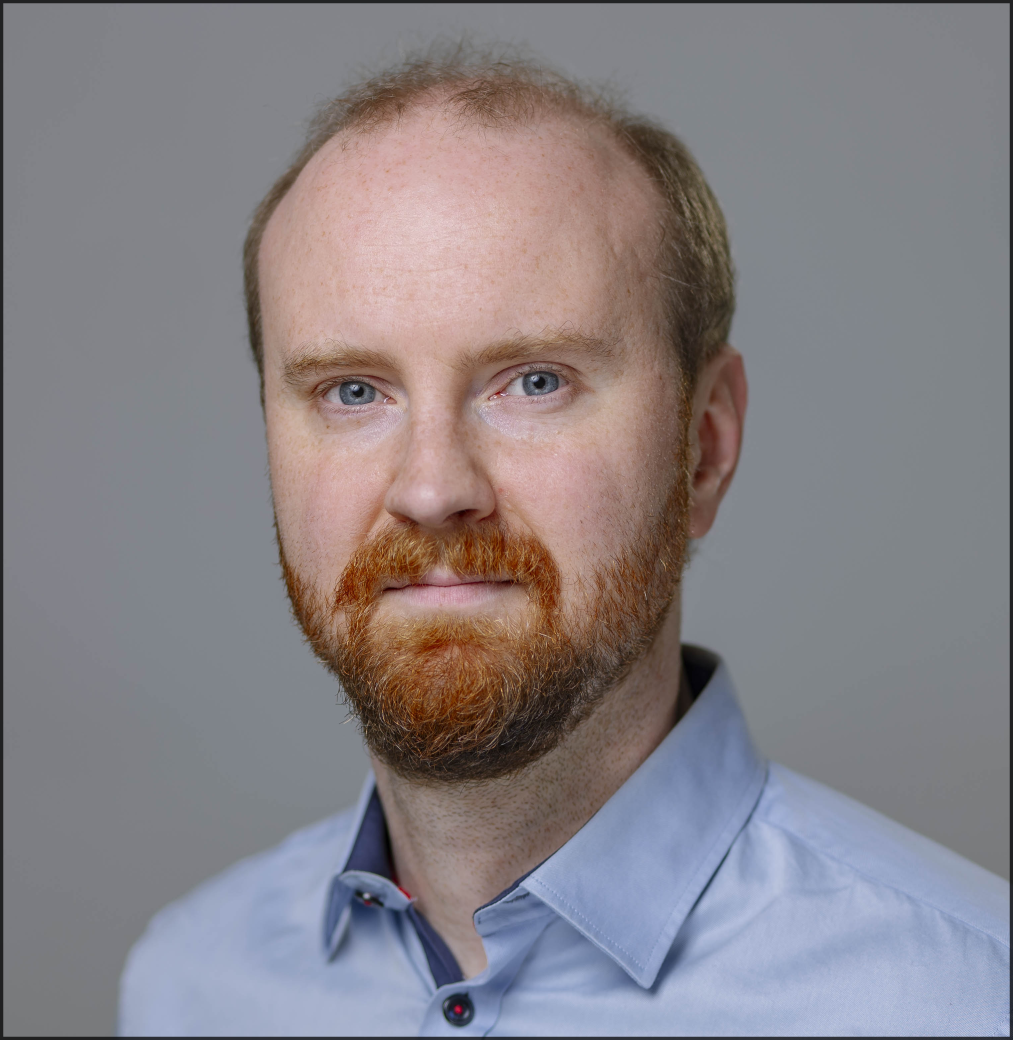}}]{Johan Wahlström} received the B.Sc., M.Sc., and Ph.D from KTH Royal Institute of Technology, Stockholm, Sweden, in 2013, 2014, and 2017, respectively. He is currently a Lecturer in Data Science with the Department of Computer Science, University of Exeter, Exeter, U.K. Between 2018 and 2020 he was a Postdoctoral Researcher with Oxford University, Oxford, U.K., working on indoor navigation for emergency responders.
\end{IEEEbiography}

\begin{IEEEbiography}[{\includegraphics[width=1in,height=1.25in,clip,keepaspectratio]{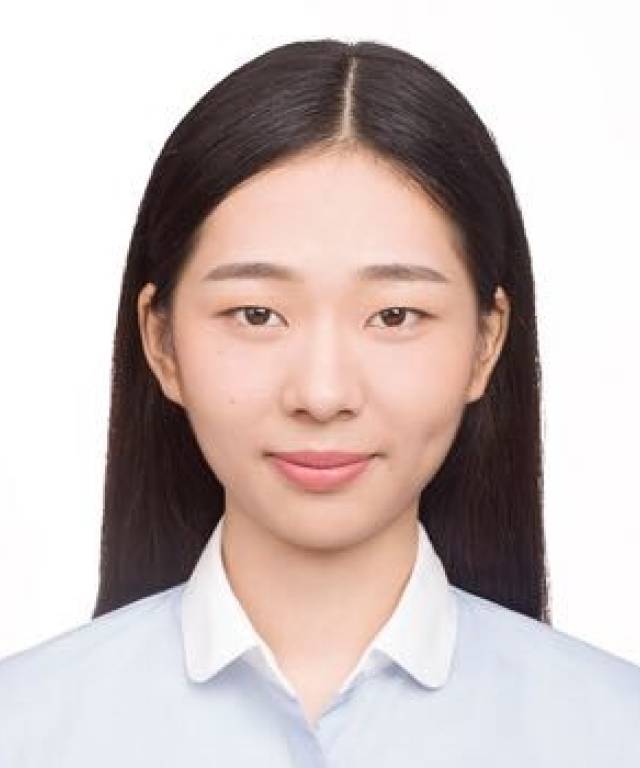}}]{Man Luo} received the B.Sc. (Hons.) and Ph.D. degrees in computer science from the University of Warwick. She was a Research Associate at the Urban Analytics Programme, The Alan Turing Institute, investigating potential policy impact in the COVID-19 pandemic and building resilience against future shocks. She is currently a Lecturer with the Department of Computer Science, University of Exeter, U.K. Her research interests include urban analytics, data systems, and spatio–temporal deep learning.
\end{IEEEbiography}

\end{document}